\documentclass{article}

\PassOptionsToPackage{square,sort,comma,numbers}{natbib}
\usepackage[preprint]{nips_2018}

\usepackage[utf8]{inputenc} 
\usepackage[T1]{fontenc}    
\usepackage{hyperref}       
\usepackage{url}            
\usepackage{booktabs}       
\usepackage{amsfonts}       
\usepackage{nicefrac}       
\usepackage{microtype}      
\usepackage{amsmath}
\usepackage{tikz}
\usepackage{subfigure}
\usepackage{pgfplots}
\usetikzlibrary{matrix}
\usepgfplotslibrary{groupplots}
\usepackage{overpic}
\usepackage{multirow}
\usepackage{appendix}

\newlength\figureheight 
\newlength\figurewidth 

\newcommand{\etal}{\mbox{\emph{et al.\ }}}

\title{PeerNets: Exploiting Peer Wisdom Against Adversarial Attacks}

\author{
  Jan Svoboda\textsuperscript{1,2}, Jonathan Masci\textsuperscript{2}, Federico Monti\textsuperscript{1,4}, Michael M. Bronstein\textsuperscript{1,3,4,5}, Leonidas Guibas\textsuperscript{6} \\
  \textsuperscript{1}USI, Switzerland \hspace{5mm}
  \textsuperscript{2}NNAISENSE, Switzerland \hspace{5mm}
  \textsuperscript{3}Intel Perceptual Computing, Israel\\
    \textsuperscript{4}Imperial College London, UK \hspace{5mm} \textsuperscript{5}Fabula AI, UK\hspace{5mm}
  \textsuperscript{6}Stanford University, USA \\
  \texttt{\{jan.svoboda,federico.monti,michael.bronstein\}@usi.ch}\\\texttt{jonathan@nnaisense.com, guibas@cs.stanford.edu} \\
}

\begin{document}

\maketitle

\begin{abstract}
Deep learning systems have become ubiquitous in many aspects of our lives. Unfortunately, it has been shown that such systems are vulnerable to adversarial attacks, making them prone to potential unlawful and harmful uses. 
Designing deep neural networks that are robust to adversarial attacks is a fundamental step in making such systems safer and deployable in a broader variety of applications (e.g., autonomous driving), 
but more importantly is a necessary step to design novel and more advanced architectures built 
on new computational paradigms rather than marginally modifying existing ones.
In this paper we introduce {\bf PeerNets}, a novel family of convolutional networks alternating classical Euclidean convolutions with graph convolutions to harness information from a graph of 
peer samples. This results in a form of non-local forward propagation in the model, where latent features are
conditioned on the global structure induced by the data graph, that is up to $3 \times$ more robust to a variety of white- and black-box adversarial attacks compared to conventional architectures
with almost no drop in accuracy.
\end{abstract}

\section{Introduction}

Deep convolutional networks (CNN) are the de-facto standard in almost any computer vision
application, ranging from image recognition~\cite{krizhevsky2012a,he2015b}, 
object detection~\cite{yuanLWYG17,redmonDGF16,renNIPS15fasterrcnn}, 
semantic segmentation~\cite{he2017maskrcnn,badrinarayanan2015segnet} 
and motion estimation~\cite{liuYTLA17,vijayanarasimhan17,niklausML17}.
The ground breaking results of deep learning for industry-relevant problems have made them also a critical ingredient in many real world systems for applications such as autonomous driving and user authentication where safety or security is important.
Unfortunately, it has been shown~\cite{szegedy2013intriguing} that such systems are vulnerable to easy-to-fabricate
adversarial examples, opening potential ways for their untraceable and unlawful exploitation.

{\bf Adversarial attacks.}
Szegedy \etal 
\cite{szegedy2013intriguing} found that deep neural networks employed in computer vision tasks tend to learn very discontinuous input-output mappings, and can be forced  to misclassify an image by applying an almost imperceptible `adversarial' perturbation to it, which is found by maximizing the network's prediction error. 
While it was expected that addition of noise to the input can degrade the classification accuracy of a neural network, the fact that only tiny adversarial perturbations are needed came as a great surprise to the machine learning community. 
Multiple methods for designing adversarial perturbations have been proposed, perhaps the most 
dramatic being single pixel perturbation \cite{suOnePixAttack2017}.
Because of its potentially serious implications in the security/safety of deep learning technologies (which are employed for example, in face recognition systems \cite{sharif2016accessorize} or autonomously driving cars), adversarial attacks and defenses against them have become a very active topic of research \cite{rauber2017,kurakin2018}. 


Adversarial attacks can be categorized as {\em{targeted}} and {\em{non-targeted}}. The former 
aims at changing a source data sample in a way to make the network classify it as some pre-specified 
target class (imagine as an illustration a villain that would like a computer vision system in a car to classify all traffic signs as STOP signs). 
The latter type of attacks, on the other hand, aims simply at making the classifier predict some different class. 
Furthermore, we can also distinguish between the ways in which the attacks are generated: {\em white box} attacks assume the full knowledge of the model and of all its 
parameters and gradient, while in {\em black box} attacks, the adversary can observe only the 
output of a given generated sample  and has no access to the model. 
%
It has been recently found that for a given model and a given dataset there 
exists a {\em universal adversarial perturbation}~\cite{Dezfooli2016} that is network-specific but input-agnostic, i.e., a single noise image that can be applied to any input of the network and causes misclassification with high-probability. Such perturbations were furthermore shown to be insensitive to geometric transformations. 


{\bf Defense against adversarial attacks.}  
There has been significant recent effort trying to understand the theoretical nature of such attacks \cite{goodfellow2015,fawzi2018analysis} and develop strategies against adversarial perturbations. 
The search for solutions has focused on the network architecture, training procedures, and data pre-processing \cite{gu2014towards}. 
Intuitively, one could argue that robustness to adversarial noise might be achieved by adding adversarial examples during training. 
However, recent work has shown that such a brute-force approach does not work in this case --- and that it forces
the network to converge to a bad local minimum~\cite{tramer2018ensemble}. As an alternative, 
\cite{tramer2017ensemble} introduced ensemble adversarial training, augmenting training data with perturbations transferred from other models.
In other efforts, \cite{papernot2016effectiveness} proposed defensive distillation training, 
while \cite{mahdizadehaghdam2018deep} showed that a deep dictionary learning architecture provides greater robustness against adversarial noise. 
Despite all these attempts, as of today, the defense against adversarial attacks is still an open research question.

{\bf Main contributions. }
In this paper, we introduce {\em Peer-Regularized networks} (PeerNet), a new family of deep models that
use a graph of samples to perform {\em non-local forward propagation}. The output of the network
depends not only on the given test sample at hand, but also on its interaction with 
several related training samples. 
We experimentally show that such a novel paradigm leads to substantially
higher robustness (up to $3 \times$) to adversarial attacks at the cost of a minor
classification accuracy drop when using the same architecture. We also show that
the proposed non-local propagation acts as a strong regularizer that makes possible
to increase the model capacity to match the current state-of-the-art without incurring
overfitting.
%
We provide experimental validation on established benchmarks such as 
MNIST, CIFAR10 and CIFAR100 using various types of adversarial attacks.





\section{Related works}

Our method is related to three classes of approaches: deep learning on graphs, deep learning models combined with manifold regularization, and non-local filtering. 
In this section, we review the related literature and provide the necessary background.

\subsection{Deep Learning on Graphs}

In the recent years, there has been a surge of interest in generalizing successful deep learning models to non-Euclidean structured data such as graphs and manifolds, a field referred to as {\em geometric deep learning} \cite{bronstein2017geometric}. 
First attempts at learning on graphs date back to the works of 
\cite{gori2005new,scarselli2009graph}, where the authors considered the steady state of a learnable diffusion process (more recent works \cite{li2016gated,gilmer2017neural} improved this approach using modern deep learning schemes).   

{\bf Spectral domain graph CNNs. }
Bruna \etal \cite{bruna2013spectral,henaff2015deep}  proposed formulating convolution-like operations in the spectral domain, defined by the eigenvectors of the graph Laplacian. 
%
%
%
%
Among the drawbacks of this architecture is $\mathcal{O}(n^2)$ computational complexity due to the cost of computing the forward and inverse graph Fourier transform, 
$\mathcal{O}(n)$ parameters per layer, and no guarantee of spatial localization of the filters.

A more efficient class of spectral graph CNNs was introduced in \cite{defferrard2016convolutional,kipf2016semi} and follow up works, who proposed spectral filters that can be expressed in terms of simple operations (such as additions, scalar- and matrix multiplications) w.r.t. the Laplacian. In particular, \cite{defferrard2016convolutional} considered polynomial filters 
%
of degree $p$, which incur 
only $p$ times multiplication by the Laplacian matrix (which costs $\mathcal{O}(| \mathcal{E} |)$ in general, or $\mathcal{O}(n)$ if the graph is sparsely connected), while also guaranteeing filters that are supported in $p$-hop neighborhoods. 
Levie \etal \cite{levie2017cayleynets} proposed rational filter functions including additional inversions of the Laplacian, which were carried out approximately using an iterative method. 
Monti \etal used multivariate polynomials w.r.t. multiple Laplacians defined by graph motifs \cite{monti2018motifnet} as well as Laplacians defined on multiple graphs \cite{monti2017geometric} in the context of matrix completion problems.

{\bf Spatial domain graph CNNs. }
On the other hand, spatial formulations of graph CNNs  operate on local neighborhoods on the graph  
\cite{duvenaud2015convolutional,monti2016geometric,atwood2016diffusion,hamilton2017inductive,velivckovic2017graph,wang2018dynamic}. 
Monti \etal \cite{monti2016geometric} proposed the Mixture Model networks (MoNet), generalizing the notion of image `patches' to graphs. The centerpiece of this construction is a system of local pseudo-coordinates $\mathbf{u}_{ij} \in \mathbb{R}^d$ assigned to a neighbor $j$ of each vertex $i$. The spatial analogue of a convolution is then defined as a Gaussian mixture in these coordinates. 
%
%
Veli\v{c}kovi\`c \etal \cite{velivckovic2017graph} reinterpreted this scheme as {\em graph attention} (GAT), learning the relevance of neighbor vertices for the filter result, 
\begin{equation} \label{eq:gat_layer}
	\tilde{\mathbf{x}}_i = \mathrm{ReLU}\left(\sum_{j \in \mathcal{N}_i} \alpha_{ij} \mathbf{x}_j\right),
\hspace{7.5mm}	\alpha_{ij} = \frac{
\exp(\mathrm{LeakyReLU}(\mathbf{b}^\top[\mathbf{A}\mathbf{x}_i,\, \mathbf{A}\mathbf{x}_j])
}
{
\sum_{k \in \mathcal{N}_i} \exp(\mathrm{LeakyReLU}(\mathbf{b}^\top[\mathbf{A}\mathbf{x}_i,\, \mathbf{A}\mathbf{x}_j] )
} \,,
\end{equation}
where $\alpha_{ij}$ are attention scores representing the importance of vertex $j$ w.r.t. $i$, 
and the $p \times q'$ matrix $\mathbf{A}$ and $2p$-dimensional vector $\mathbf{b}$ are the learnable parameters.

\subsection{Low-Dimensional Regularization}
There have recently been several attempts to marry deep learning with classical manifold learning methods \cite{belkin2003laplacian,coifman2006diffusion} based on the assumption of local regularity of the data, that can be modeled as a low-dimensional manifold in the data space. 
\cite{zhu2017ldmnet} proposed a feature regularization method that makes input and output
features to lie on a low dimensional
manifold. 
 \cite{garcia2017few} cast few shot learning as supervised message passing task which is trained end-to-end using graph neural networks.
One of the key disadvantages of such approaches is the difficulty of factoring out global transformations (such as translations) to which the network output should be invariant, but which ruin local similarity.

\subsection{Non-Local Image Filtering}
The last class of methods related to our approach are 
non-local filters~\cite{sochen1998general,tomasi1998bilateral,Buades2005} that gained popularity in the image processing community about two decades ago. Such non-shift-invariant filters produce, for every location of the image, a result that depends not only on the pixel intensities
around the point, but also on a set of neighboring pixels and their relative locations. 
For example, the bilateral filter uses radiometric differences together with 
Euclidean distance of pixels to determine the averaging weights.
The same paradigm was brought into non-local networks~\cite{wang2017non}, 
where the response at a given position is defined as a weighted sum of all the features across different spatio-temporal locations,
whose interpolation coefficients are inferred by a parametric model, usually another neural network.

\section{Peer Regularization}




In this paper, we propose a new deep neural network architecture that takes advantage of the data space structure. 
The centerpiece of our model is the {\em Peer Regularization} (PR) layer, designed as follows. 
Let $\mathbf{X}^1, \hdots, \mathbf{X}^N$ be $n\times d$ matrices representing the feature maps of $N$ images, to which we refer as {\em peers} (here $n$ denotes the number of pixels and $d$ is the dimension of the feature in each pixel). 
Given a pixel of image $i$, we consider its $K$ nearest neighbor graph in the space of $d$-dimensional feature maps of all pixels of all the peer images, where the neighbors are computed using e.g. the cosine distance. The $k$th nearest neighbor of the $p$th pixel $\mathbf{x}^i_{p}$ taken from image $i$ is the $q_k$th pixel $\mathbf{x}^{j_k}_{q_k}$ taken from peer image $j_k$, with $k=1,\hdots, K$ and $j_k \in \{1,\hdots, N \}$, $q_k \in \{1,\hdots, n \}$.

We apply a variant of graph attention network (GAT) \cite{velivckovic2017graph} to the nearest-neighbor graph constructed this way, 
\begin{equation}
\label{eq:PR_layer}
\tilde{\mathbf{x}}^{i}_{p} = \sum_{k=1}^K \alpha_{i j_k p q_k} \mathbf{x}^{j_k}_{q_k}, \hspace{7.5mm} 
\alpha_{i j_k p q_k} = \frac{\mathrm{LeakyReLU}(\exp(a(\mathbf{x}^i_{p},\mathbf{x}^{j_k}_{p_k})))
}{
\sum_{k'=1}^K \mathrm{LeakyReLU}(\exp(a( \mathbf{x}^i_{p},\mathbf{x}^{j_{k'}}_{p_{k'}} )))
}\,,
\end{equation}
where $a()$ denotes a fully connected layer mapping from $2d$-dimensional input to scalar output, and $\alpha_{i j_k p q_k}$ are attention scores determining the importance of contribution of the $q_k$th pixel of image $j$ to the output $p$th pixel $\tilde{\mathbf{x}}_p^i$ of image $i$. This way, the output feature map $\tilde{\mathbf{X}}^i$ is pixel-wise weighted aggregate of the peers. 
Peer Regularization is reminiscent of non-local means denoising \cite{Buades2005}, with the important difference that the neighbors are taken from multiple images rather than from the same image, and the  combination weights are learnable. 


{\bf Randomized approximation.} In principle, one would use a graph built out of a nearest neighbor search 
across all the available training samples.  
Unfortunately, this is not feasible due to memory and computation limitations. We therefore use a Monte Carlo approximation, as follows. 
%
Let $\mathbf{X}^1, \hdots, \mathbf{X}^{N'}$ denote the images of the training set. We select randomly with uniform distribution $M$ smaller batches of $N\ll N'$ peers, batch $m$ containing images $\{l_{m1}, \hdots, l_{mN}\} \subset \{1,\hdots, N' \}$. 
The nearest-neighbor graph is constructed separately for each batch $m$, so that the $k$th nearest neighbor of pixel $p$ in image $i$ is pixel $j_{mk}$ in image $j_{mk}$, where $m=1,\hdots, M$, $j_{mk} \in \{l_{m1}, \hdots, l_{mN}\}$, and $p_{mk} \in \{1,\hdots, n\}$.  
%
%
The output of the filter is approximated by empirical expectation on the $M$ batches, 
estimated as follows
\begin{equation} 
\tilde{\mathbf{x}}^{i}_{p} = \frac{1}{M} \sum_{m = 1}^{M} \sum_{k=1}^K \alpha_{i j_{mk} p q_{mk}} \mathbf{x}^{j_{mk}}_{q_{mk}}. 
\end{equation}
In order to limit the computational overhead, $M=1$ is used during training, whereas larger values of $M$ are used during
inference (see Section~\ref{sec:experiments} for details). 

\begin{figure}[ht!]
\centering
\includegraphics[scale=0.275]{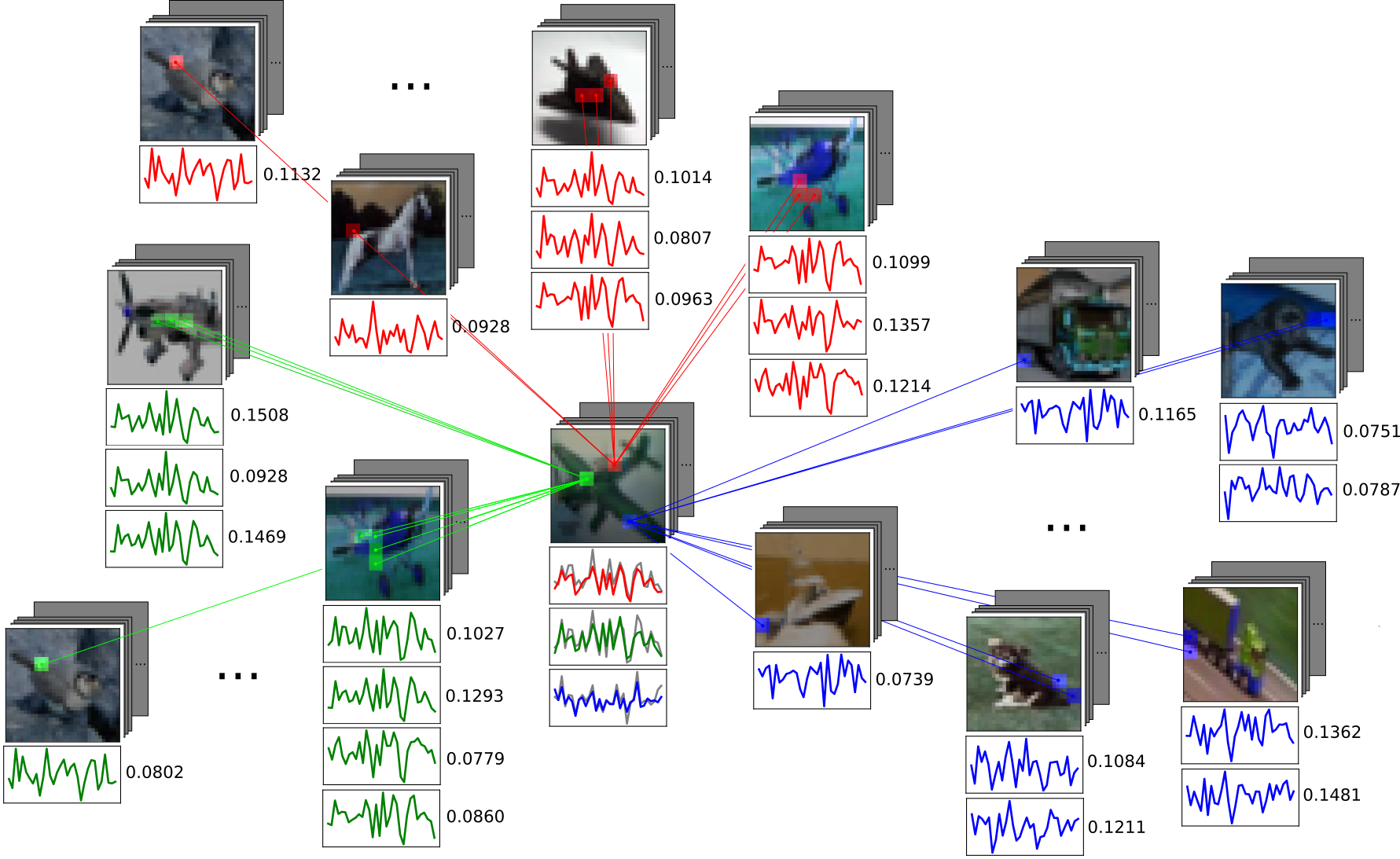}
\caption{ Our Peer Regularization illustrated on three pixels (red, green, blue) of a CIFAR image (center). For each pixel, $K$ nearest neighbors are found in peer images. Plots represent the feature maps in the respective pixels; numbers represent the attention scores. }
\label{fig:peerRegularizationScheme}
\end{figure}

\section{Experiments}
\label{sec:experiments}
We evaluate the robustness of Peer Regularization to adversarial perturbations on standard benchmarks (MNIST \cite{lecun1998mnist}, CIFAR-10, and CIFAR-100 \cite{krizhevsky2009learning}) using a selection of common architectures (LeNet \cite{lecun1998gradient} and ResNet \cite{he2015b}). 
The modifications of the aforementioned architectures with the additional PR layers are referred to as PR-LeNet and PR-ResNet and depicted in Figure~\ref{fig:architectures}. 
%

Additional details, including hyper-parameters used during the training of models we have used throughout the experiments, are listed in Table \ref{tab:tableOptimization}.

\begin{table}[ht]
\begin{small}
\caption{Optimization parameters for different architectures and datasets. Learning rate is decreased
at epochs 100, 175, and 250 with a step factor of $10^{-1}$.}
\begin{center}
\begin{tabular}{lccccccc}
\toprule
Model & Optimizer & Epochs & Batch & Momentum & LR & $L_2$ reg. & LR decay \\
\midrule
LeNet-5 & Adam & 100 & 128 & --- & $10^{-3}$ & $10^{-4}$ & --- \\
PR-LeNet-5 & Adam & 100 & 32 & --- & $10^{-3}$ & $10^{-4}$ & --- \\
ResNet-32 & Momentum & 350 & 128 & 0.9 & $10^{-1}$ & $10^{-3}$ & step \\
PR-ResNet-32 & Momentum & 350 & 64 & 0.9 & $10^{-2}$ & $10^{-3}$ & step \\
ResNet-110 & Momentum & 350 & 128 & 0.9 & $10^{-1}$ & $2\times 10^{-3}$ & step \\
PR-ResNet-110 & Momentum & 350 & 64 & 0.9 & $10^{-2}$ & $2\times 10^{-3}$ & step \\
\bottomrule
\label{tab:tableOptimization}
\end{tabular}
\end{center}
\end{small}
\vspace*{-3mm}
\end{table}

\subsection{Graph construction}
\paragraph{Training.} During training, the graph is constructed using all the images of the current batch as peers. We compute all-pairs distances and select $K$-nearest neighbors. 
To mitigate the influence of the small batch sizes during training of PR-Nets, we introduce high level of stochasticity inside the PR layer, by using dropout of $0.2$ on the all-pairs distances while performing $K$-nearest neighbors and $0.5$ on the attention weights right before the \textit{softmax} nonlinearity. We used $K=10$ for all the experiments. 

\paragraph{Testing.} For testing, we select $N$ fixed peer images 
and then compute distances between each testing sample and all the $N$ peers. Feeding a batch of test samples, each test sample can be adjacent only to the $N$ samples in the fixed graph, and not to other test samples.
Because of the stochasticity in the graph construction and random uniform sampling of the graph nodes in the testing phase, we perform a Monte Carlo sampling over $M$ forward passes with different uniformly sampled graphs and average over $M$ runs. 
We used $N=50$ for MNIST and CIFAR-10 and $N=500$ for CIFAR-100.

\subsection{Architectures}
\begin{figure}[ht!]
\centering
\includegraphics[scale=0.60]{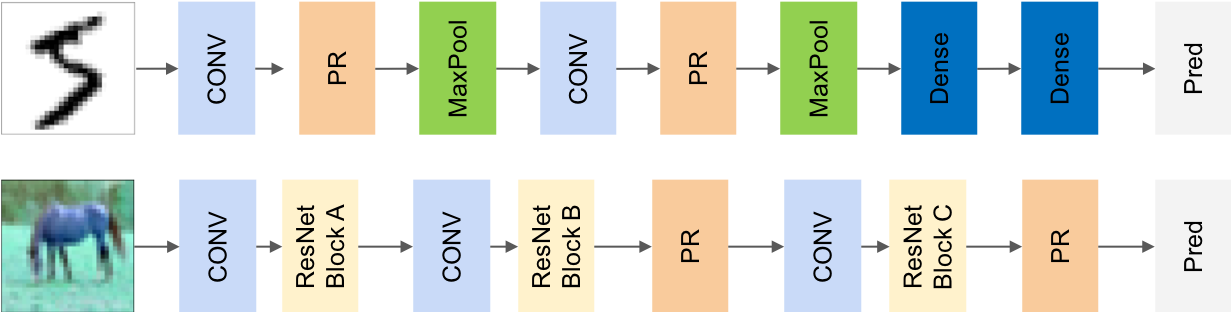} 
\caption{ {\em Top:} PR-LeNet architecture used on the MNIST dataset;
{\em Bottom:} PR-ResNet architecture used for the CIFAR experiments, 
letters A, B, and C indicate the number of feature maps in each residual block.
The respective baseline models are produced by simply removing the PR layers.}
\label{fig:architectures}
\end{figure}


{\bf MNIST.} We modify the LeNet-5 architecture by adding two PR layers, 
after each convolutional layer and before max-pooling (Figure~\ref{fig:architectures}, top). 

{\bf CIFAR-10.} We modify the ResNet-32 model, with $A=16$, $B=32$ and $C=64$, 
as depicted in Fig.~\ref{fig:architectures} by adding two PR layers at the last 
change of dimensionality and before the classifier. Each ResNet block is a sequence of Conv + BatchNorm + ReLU + Conv + BatchNorm + ReLU layers.

{\bf CIFAR-100.} For this dataset we take ResNet-110 ($A=16$, $B=32$ and $C=64$) and modify it 
in the same way as for CIFAR-10. Each block is of size 18.




\begin{figure}[ht!]
\centering
\addtocounter{subfigure}{-1}
\subfigure
{
\subfigure
{
\includegraphics[scale=2.25]{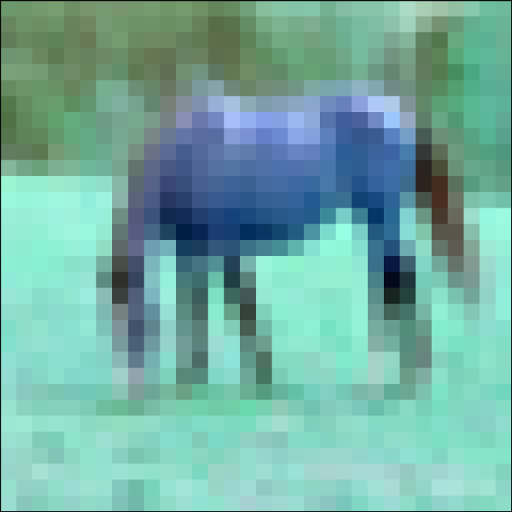}
}
\subfigure
{
\includegraphics[scale=2.25]{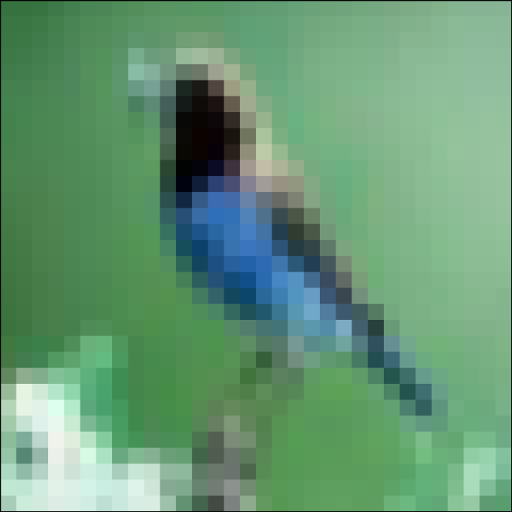}
}
\subfigure
{
\includegraphics[scale=2.25]{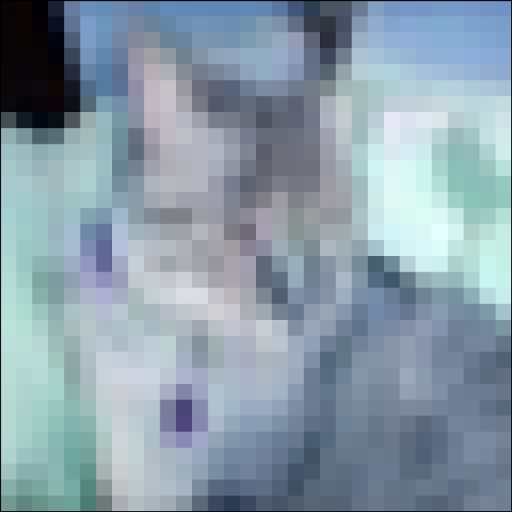}
}
\subfigure
{
\includegraphics[scale=2.25]{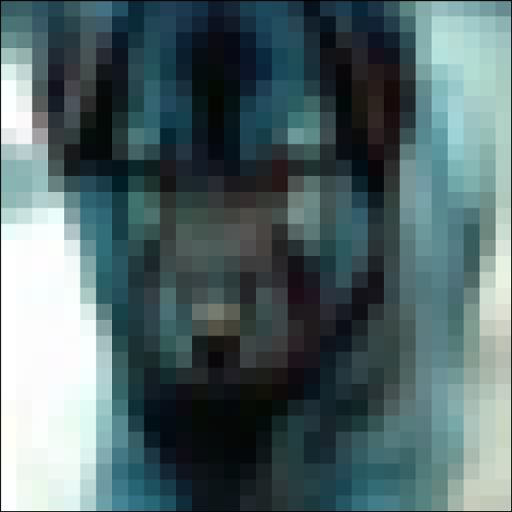}
}
\subfigure
{
\includegraphics[scale=2.25]{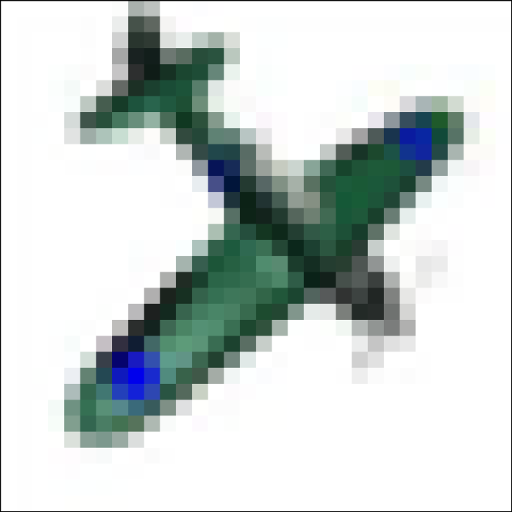}
}
}\vspace{-3mm}
\addtocounter{subfigure}{-1}
\subfigure
{
\subfigure
{
\includegraphics[scale=2.25]{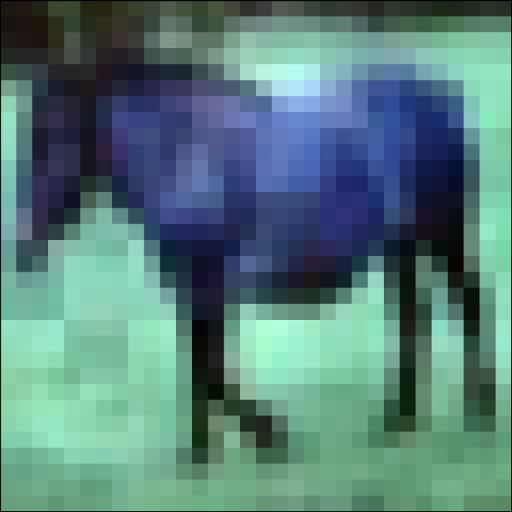}
}
\subfigure
{
\includegraphics[scale=2.25]{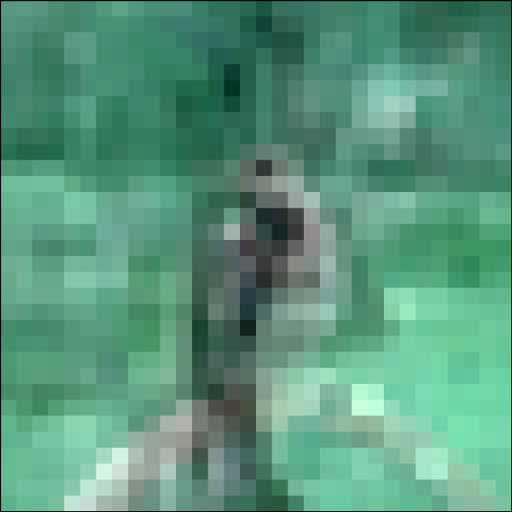}
}
\subfigure
{
\includegraphics[scale=2.25]{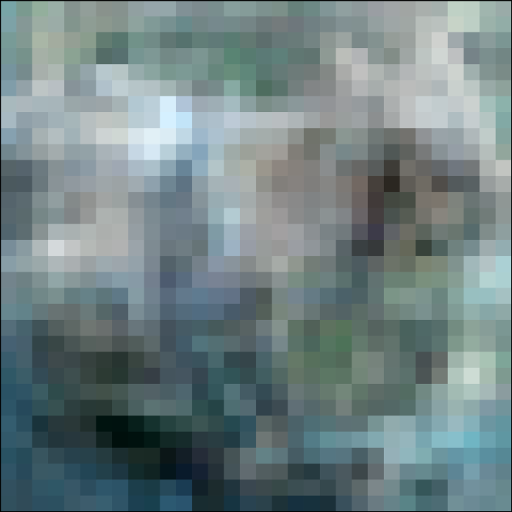}
}
\subfigure
{
\includegraphics[scale=2.25]{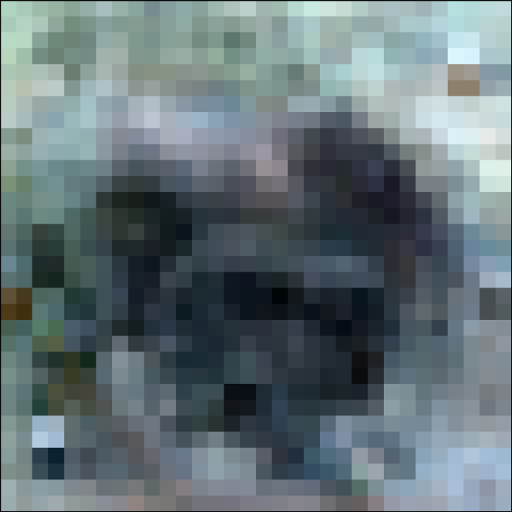}
}
\subfigure
{
\includegraphics[scale=2.25]{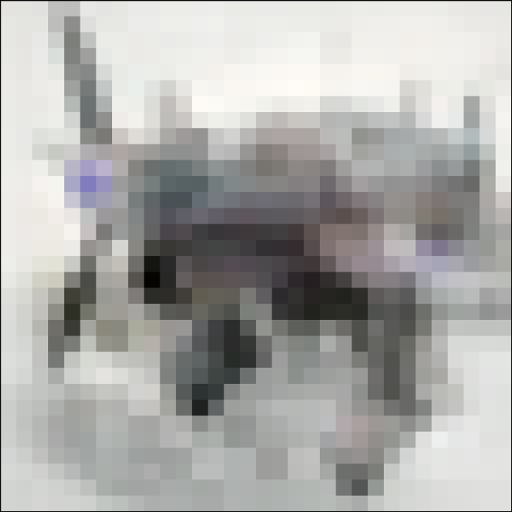}
}
}
\caption{  Examples of ``Franken-images'' (second row) constructed by backpropagating the attention scores 
to the input and using them to compose a new image as weighted sum of the peer pixels. 
Original images are shown in the first row.}
\label{fig:frankenImages}
\end{figure}

\begin{figure}[ht!]
\centering
\addtocounter{subfigure}{-1}
\subfigure
{
\subfigure
{
\includegraphics[scale=0.26]{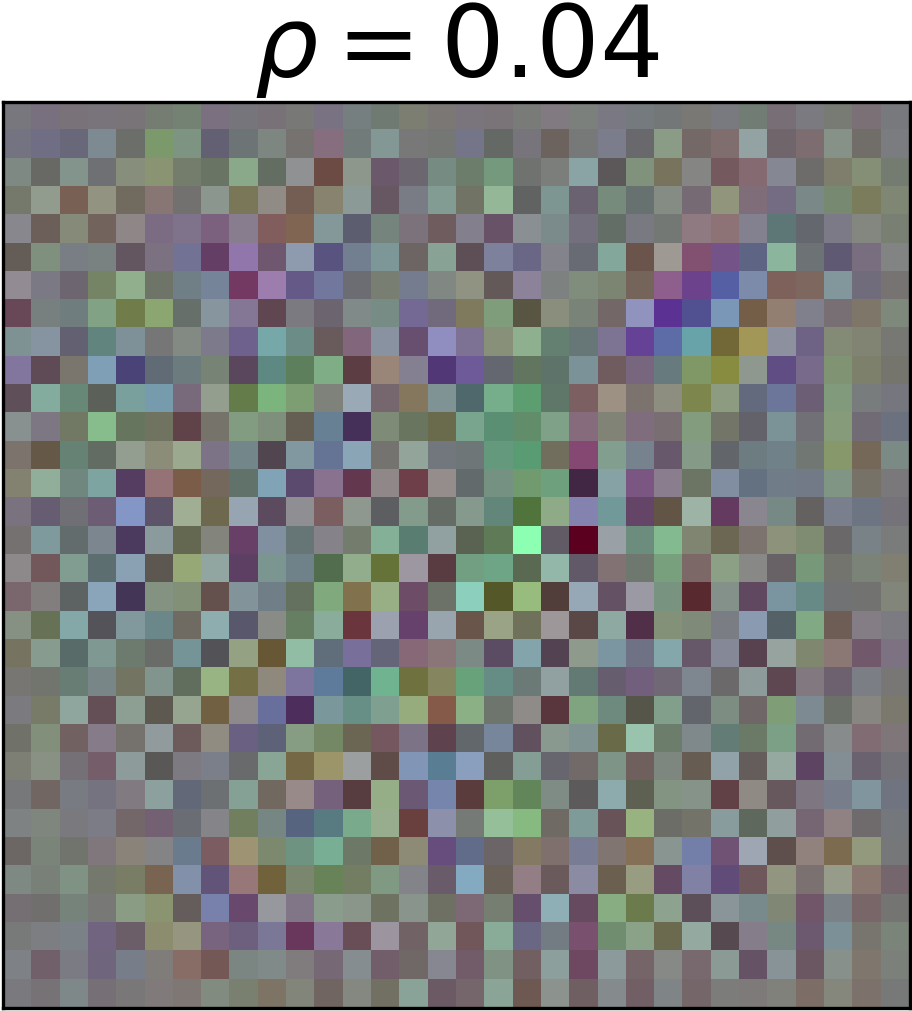}
}
\subfigure
{
\includegraphics[scale=0.26]{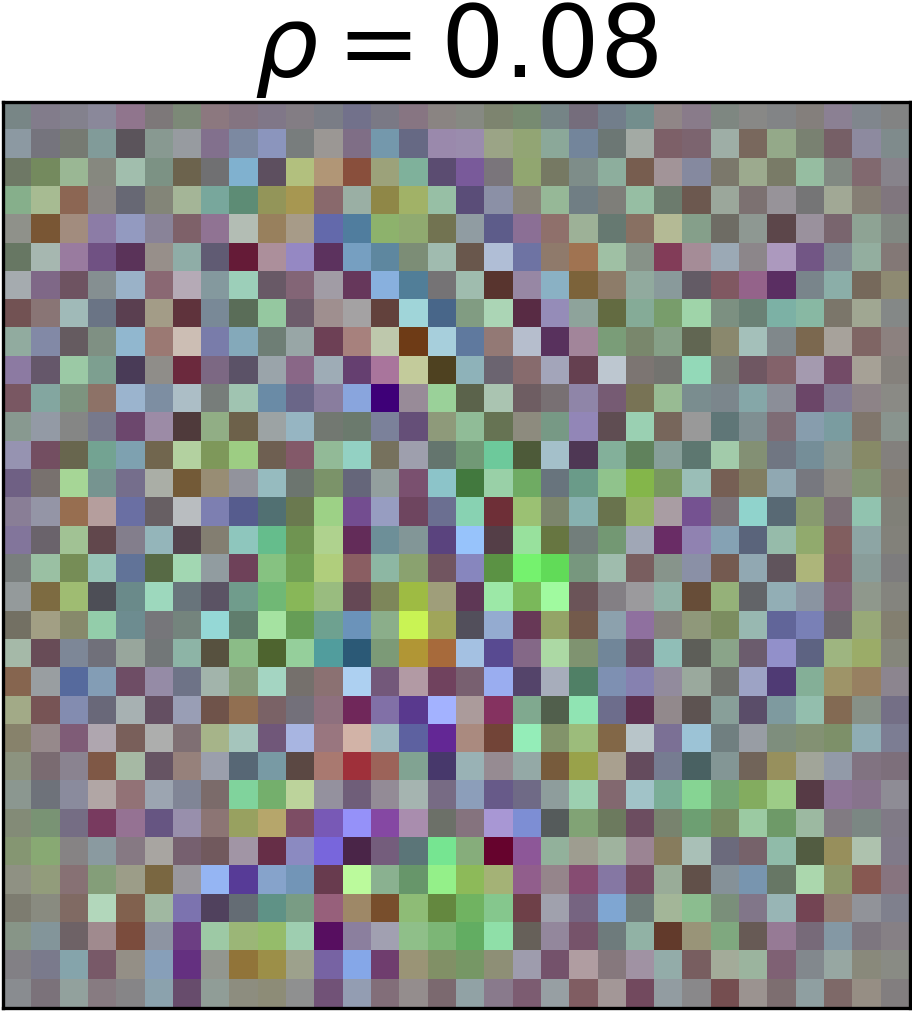}
}
\subfigure
{
\includegraphics[scale=0.26]{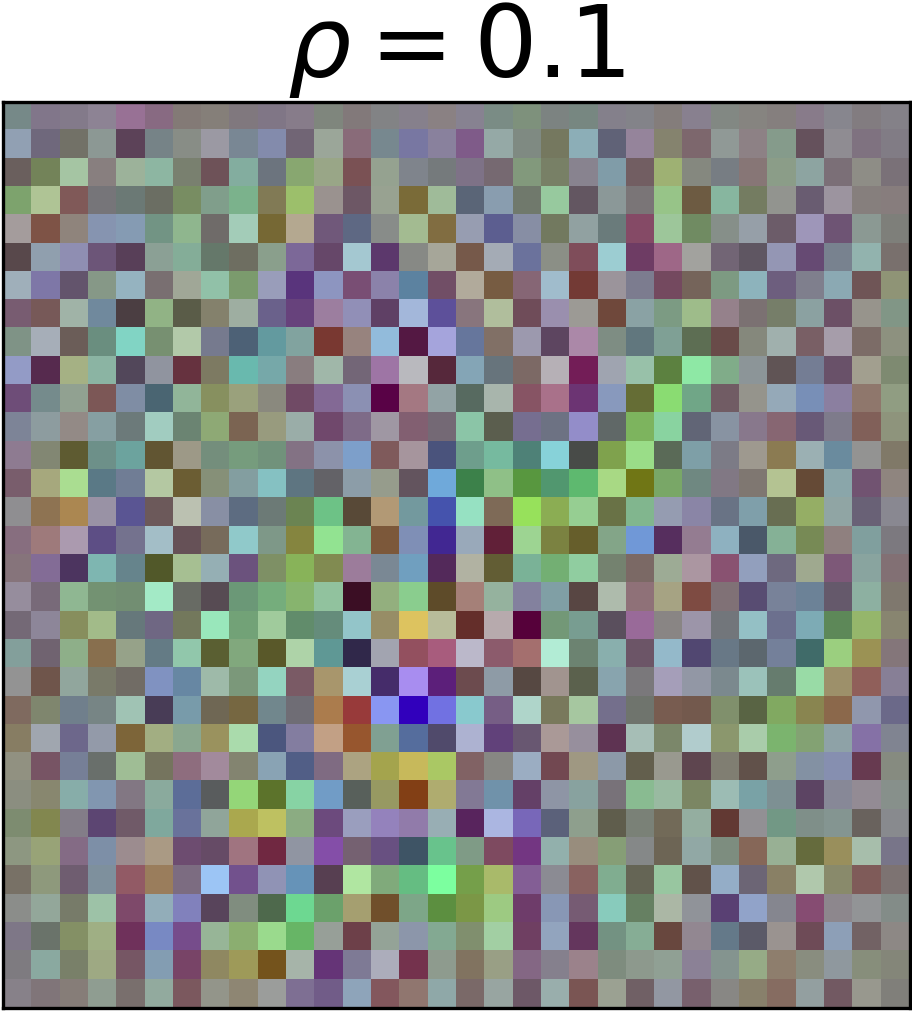}
}
\subfigure
{
\includegraphics[scale=0.26]{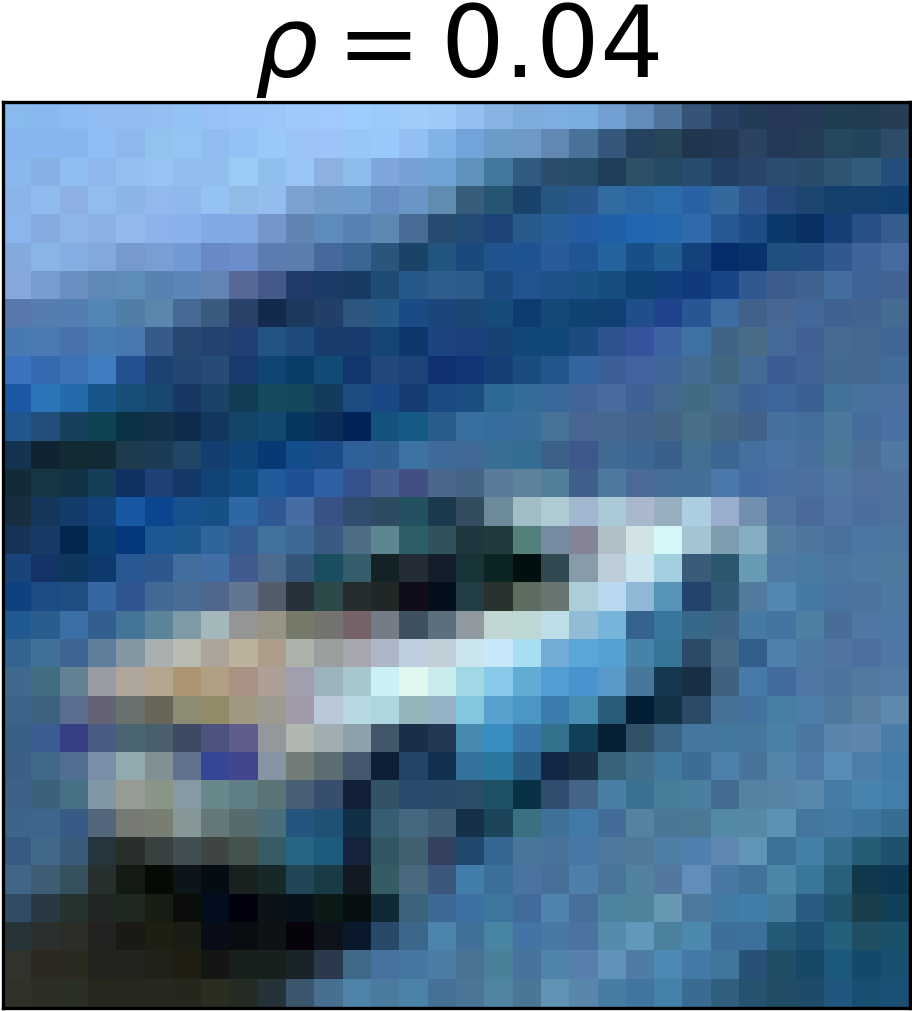}
}
\subfigure
{
\includegraphics[scale=0.26]{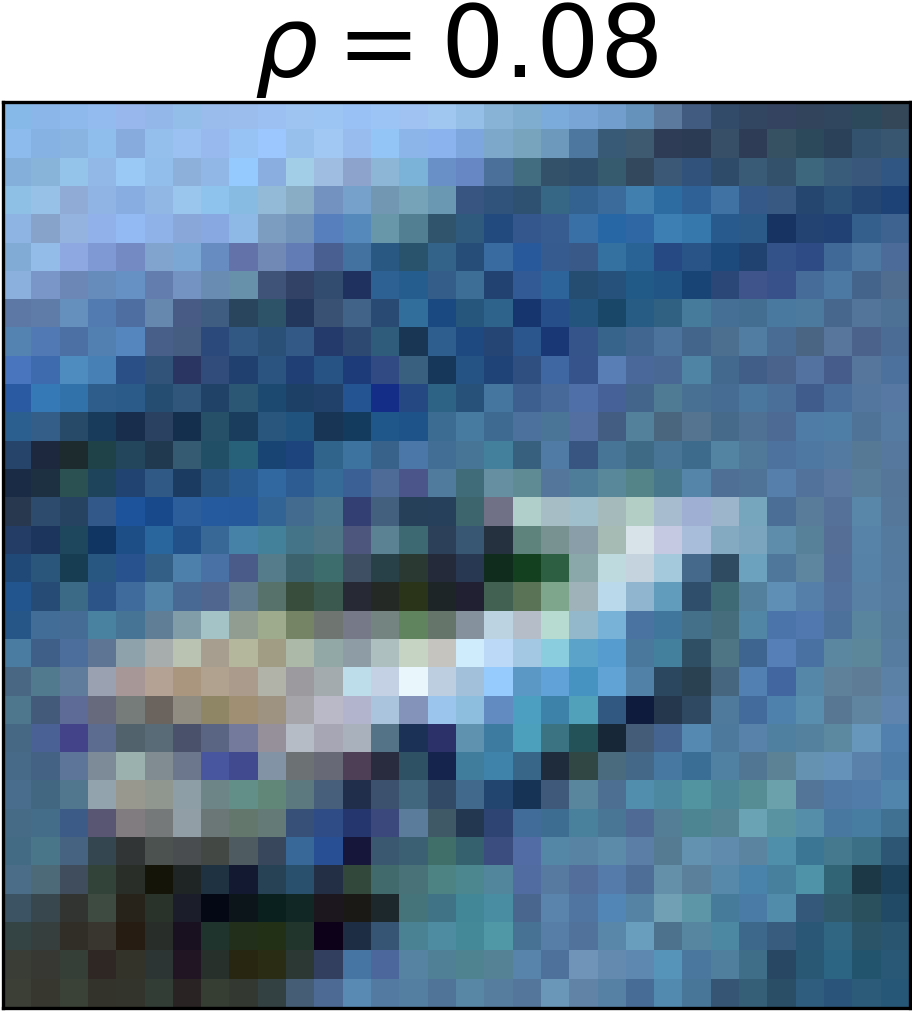}
}
\subfigure
{
\includegraphics[scale=0.26]{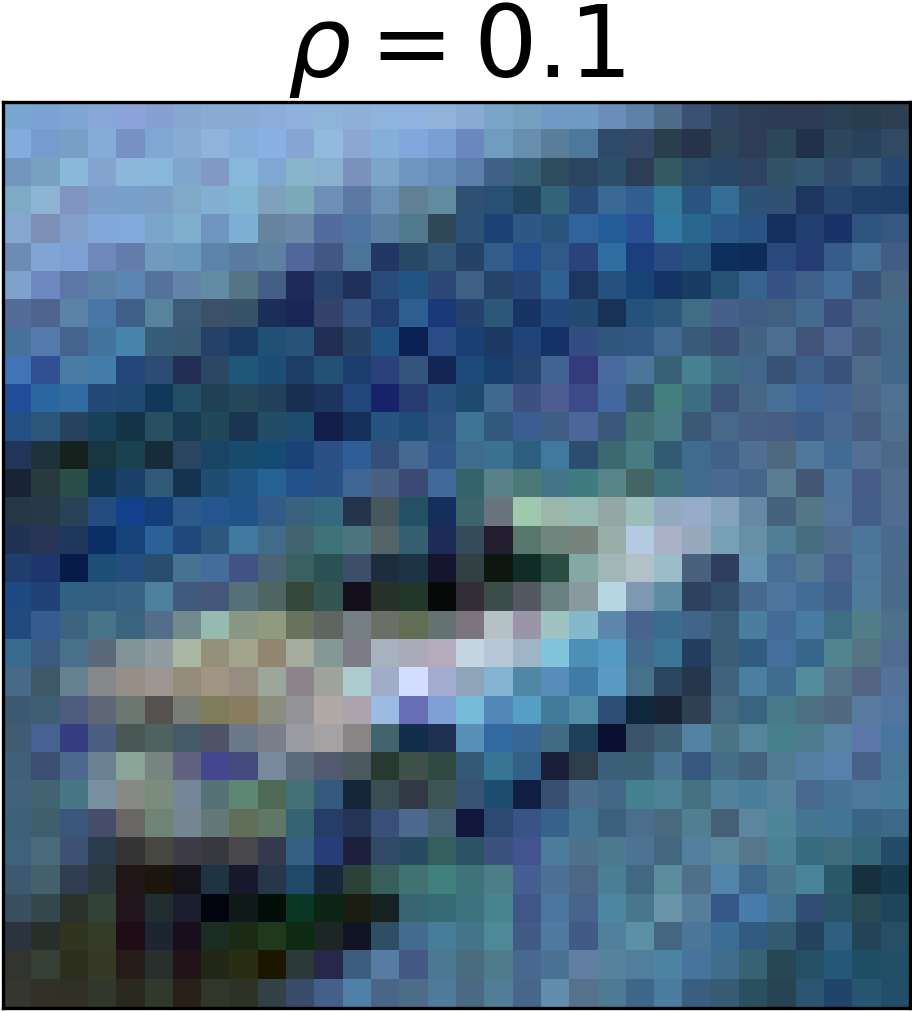}
}
}\vspace{-3mm}
\addtocounter{subfigure}{-1}
\subfigure
{
\subfigure
{
\includegraphics[scale=0.26]{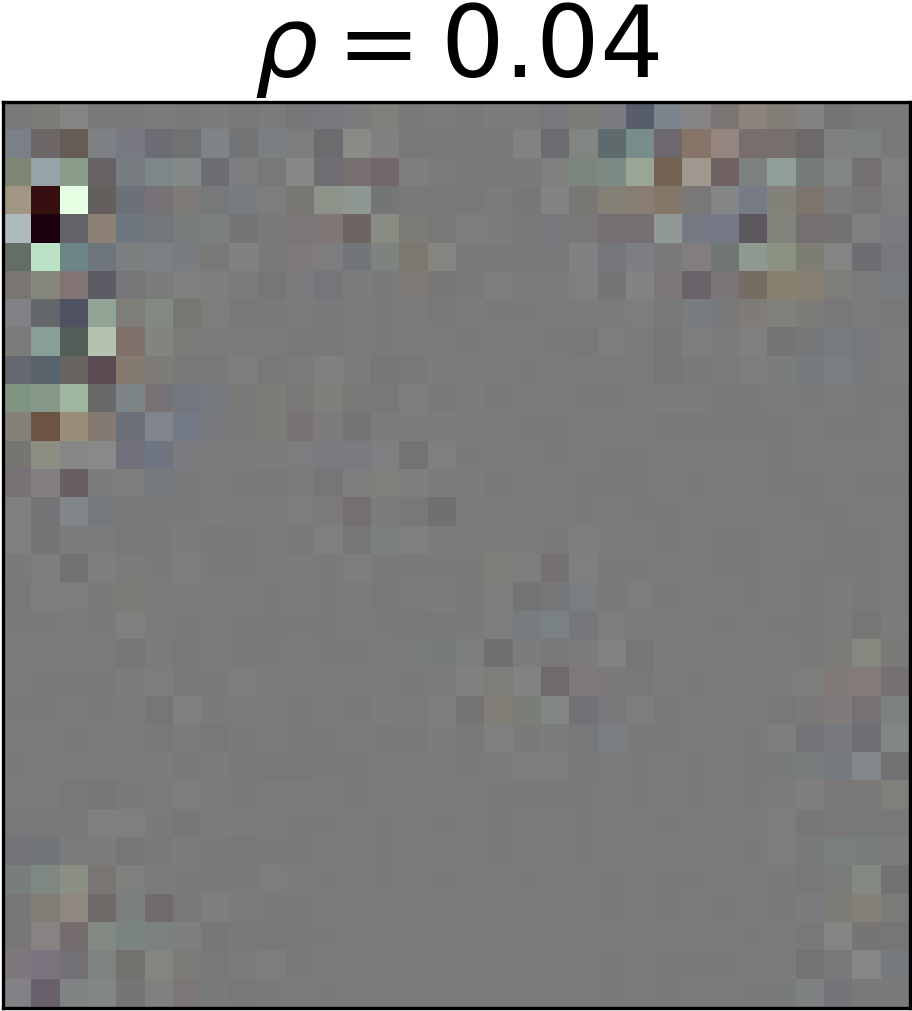}
}
\subfigure
{
\includegraphics[scale=0.26]{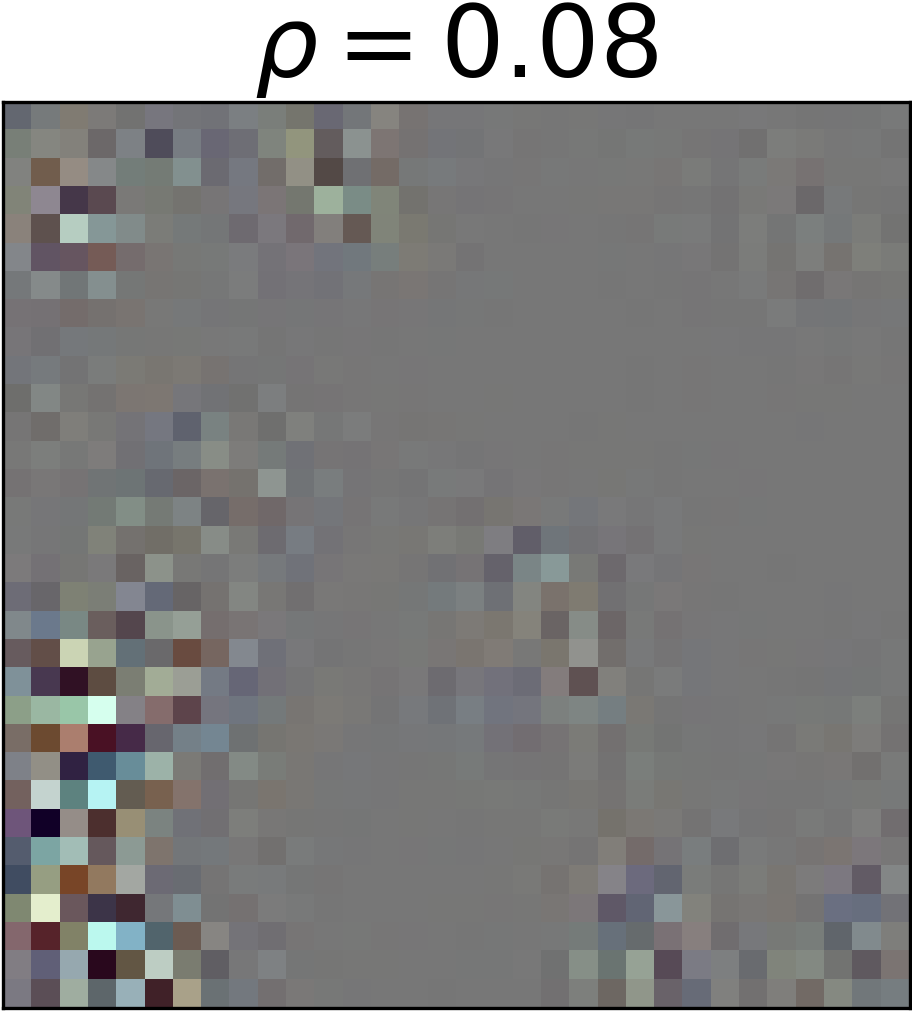}
}
\subfigure
{
\includegraphics[scale=0.26]{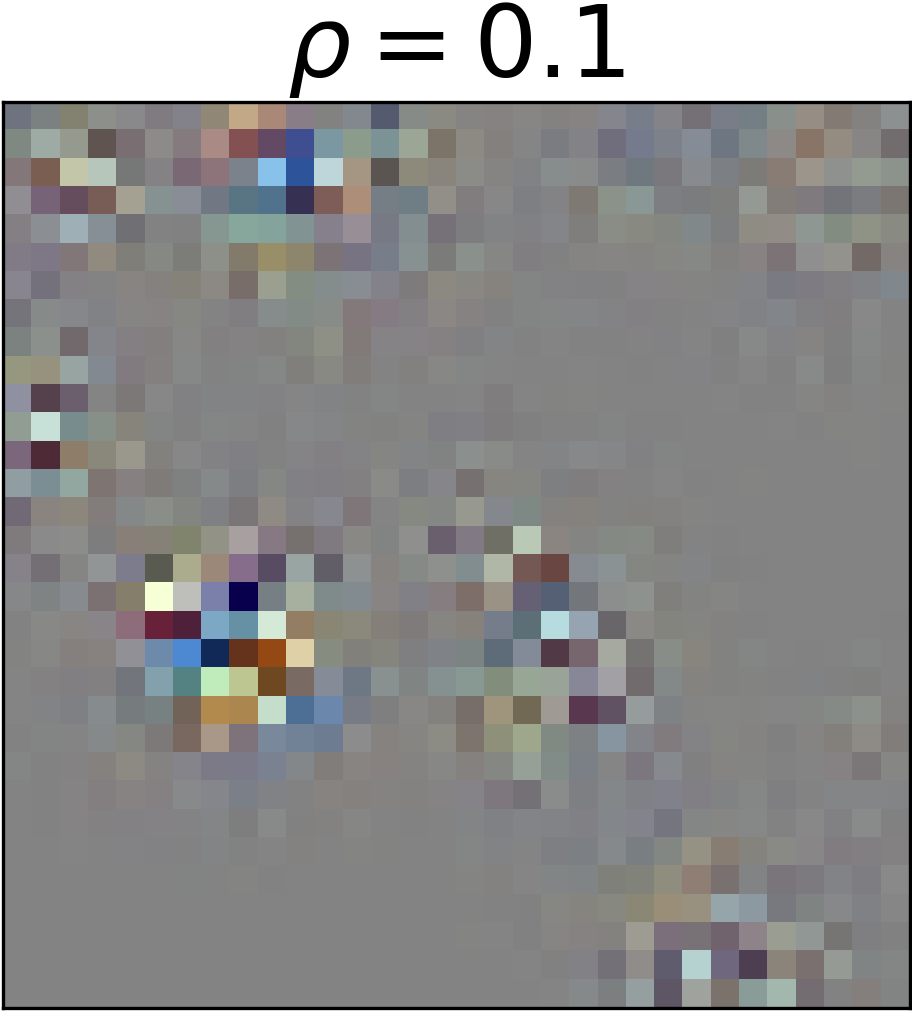}
}
\subfigure
{
\includegraphics[scale=0.26]{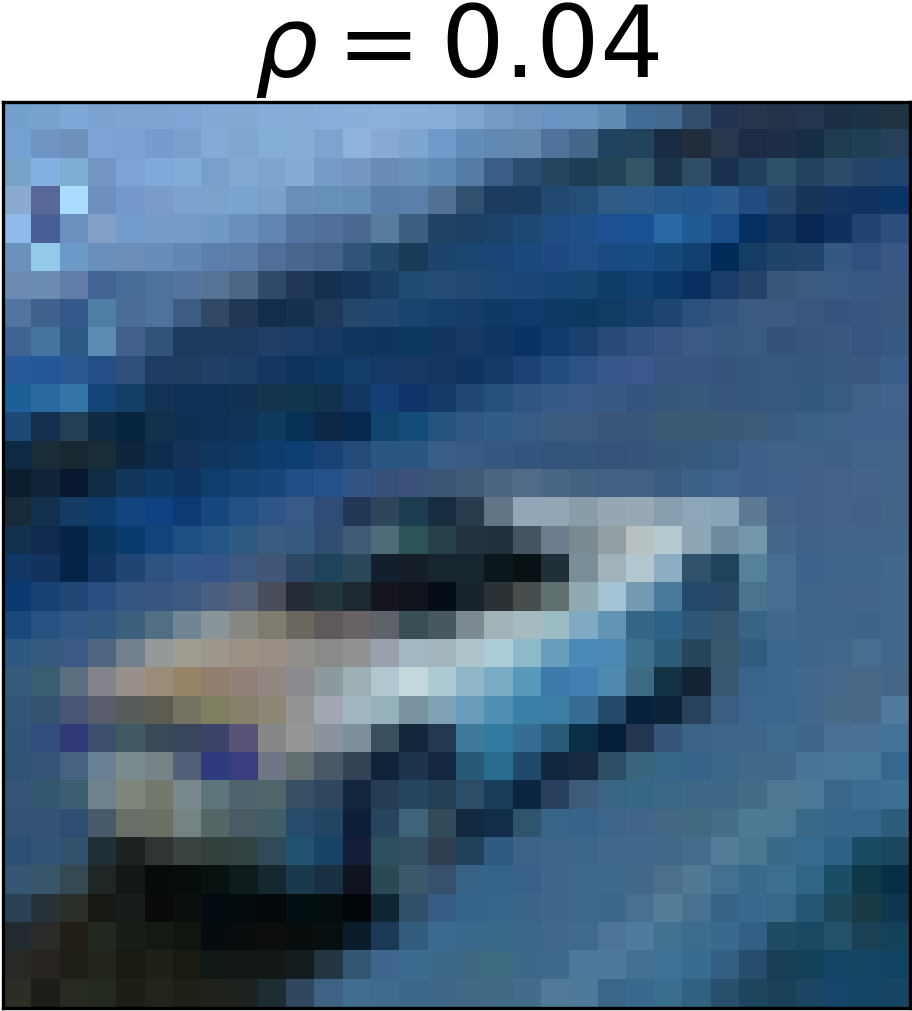}
}
\subfigure
{
\includegraphics[scale=0.26]{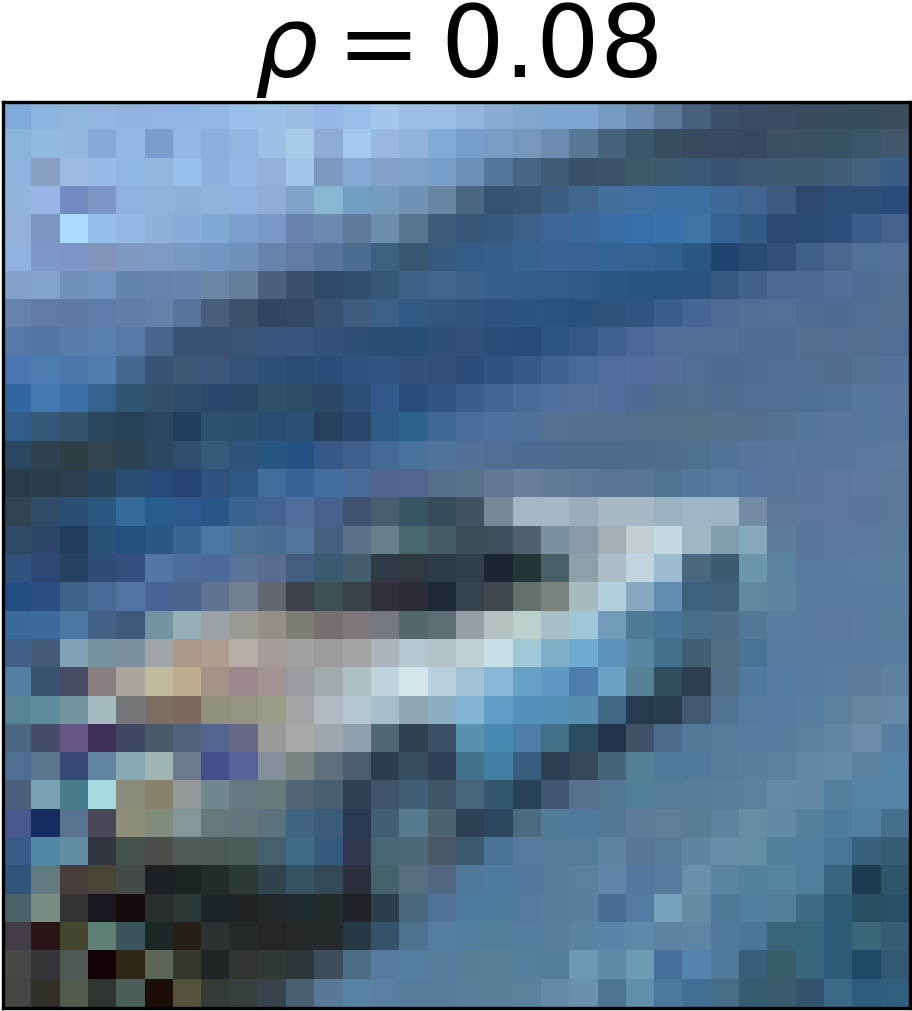}
}
\subfigure
{
\includegraphics[scale=0.26]{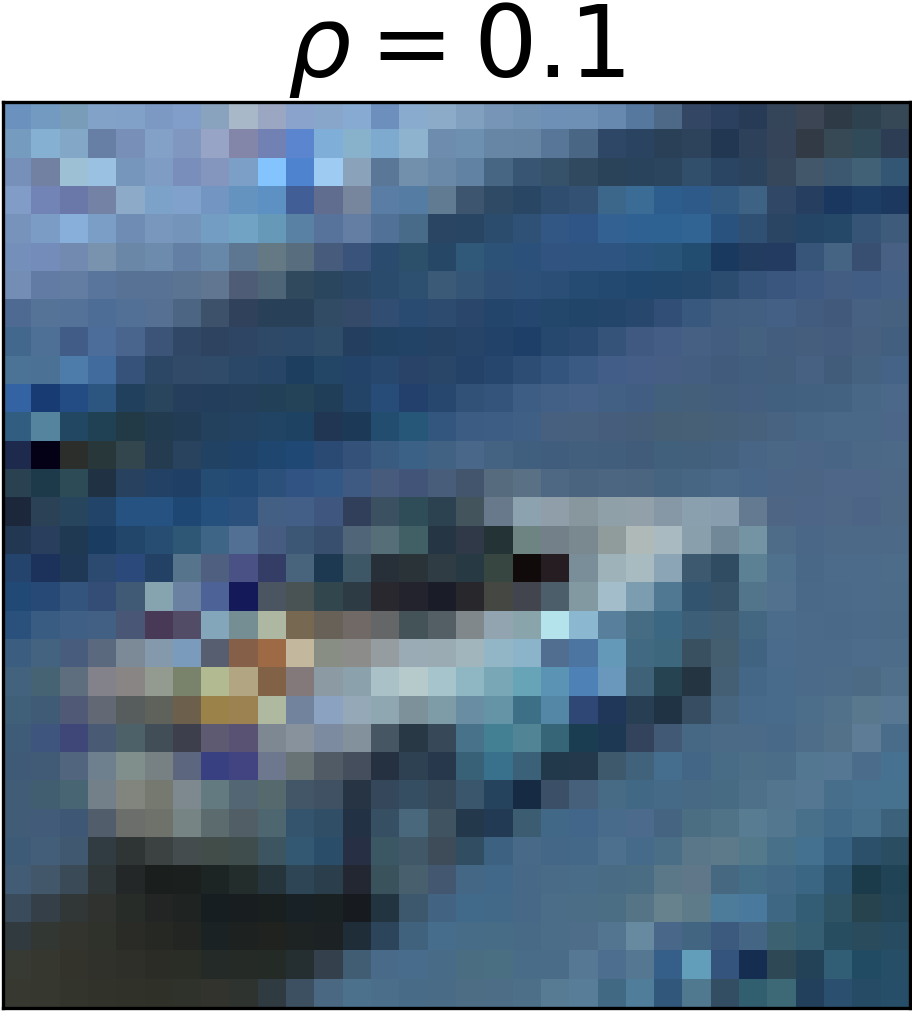}
}
}
\caption{  Examples of generated universal adversarial perturbations for CIFAR-10 dataset (left) and their applications to a sample image (right) for different values of $\rho$. Top row are the results for ResNet-32, where perturbing ``automobile'' with with $\rho=0.08$ will turn it into ``cat'' and with $\rho=0.10$ into ``bird''. The bottom row presents the results for PR-ResNet-32, where even the strong perturbation with $\rho=0.10$ will not change the classification result.}
\label{fig:perturbationUniversal}
\end{figure}

\subsection{Results}

We tested robustness against five different types of white-box adversarial attacks:
gradient based, fast-gradient sign method~\cite{goodfellow2015},
and universal adversarial perturbation\cite{Dezfooli2015}.
Similarly to previous works on adversarial attacks, we evaluate robustness 
in terms of the \textit{fooling rate}, defined as the ratio of images for 
which the network predicts a different label 
as the result of the perturbation.
As suggested in~\cite{Dezfooli2016}, random perturbations aim to shift the data points to the decision boundaries
of the classifiers. 
For the method in~\cite{Dezfooli2016}, even if the perturbations should be universal across different models, 
we generated it 
for each model separately to achieve the fairest comparison.

\paragraph{Timing.} On CIFAR-10 our unoptimized PR-ResNet-32 implementation,
during training, processes a batch of $64$ samples in $0.15s$ 
compared to the ResNet-32 baseline that takes $0.07s$. At inference time 
a batch of $100$ samples with a graph of size $100$ is processed
in $1.5s$ whereas the baseline takes $0.4s$. However, it should
be noted that much can be done to speed-up the PR layer, we leave it for future
work.

\subsubsection{Gradient descent attacks}
\label{sec:gradattack}
Given a trained neural network and an input sample $\mathbf{x}$, with conditional probability 
over the labels $p(y| \mathbf{x})$, we aim to generate an adversarial sample $\hat{\mathbf{x}}$ 
such that $p(\hat{y}| \hat{\mathbf{x}})$ is maximum for $\hat{y} \ne y$, 
while satisfying $\vert\vert \mathbf{x} - \hat{\mathbf{x}} \vert\vert_{\infty} \leq \epsilon$, 
where $\epsilon$ is some small value to produce a perturbation $\mathbf{v}$
that is nearly imperceivable.  
Generation of the adversarial example is posed as a constrained optimization problem and 
solved using gradient descent, for best reproducibility we used the implementation 
provided in Foolbox~\cite{rauber2017}. 
It can be summarized as follows:
\begin{align*}
\hat{\mathbf{x}} & \leftarrow \hat{\mathbf{x}} + \epsilon \cdot \nabla \log p(\hat{y} | \hat{\mathbf{x}}) 
\\
\hat{\mathbf{x}} & \leftarrow \mathrm{clip}(\hat{\mathbf{x}}, \mathbf{x} - \epsilon, \mathbf{x} + \epsilon)
\end{align*}
where ten $\epsilon$ values are scanned in ascending order, until the algorithm either succeeds or fails. 
All the results are reported in terms of fooling rate, in \%, and 
$\mathbb{E} \vert\vert \mathbf{v} \vert\vert_2$.

\begin{figure}[ht!]
\centering
\subfigure[Targeted (gradient descent)]
{
\includegraphics[scale=0.225]{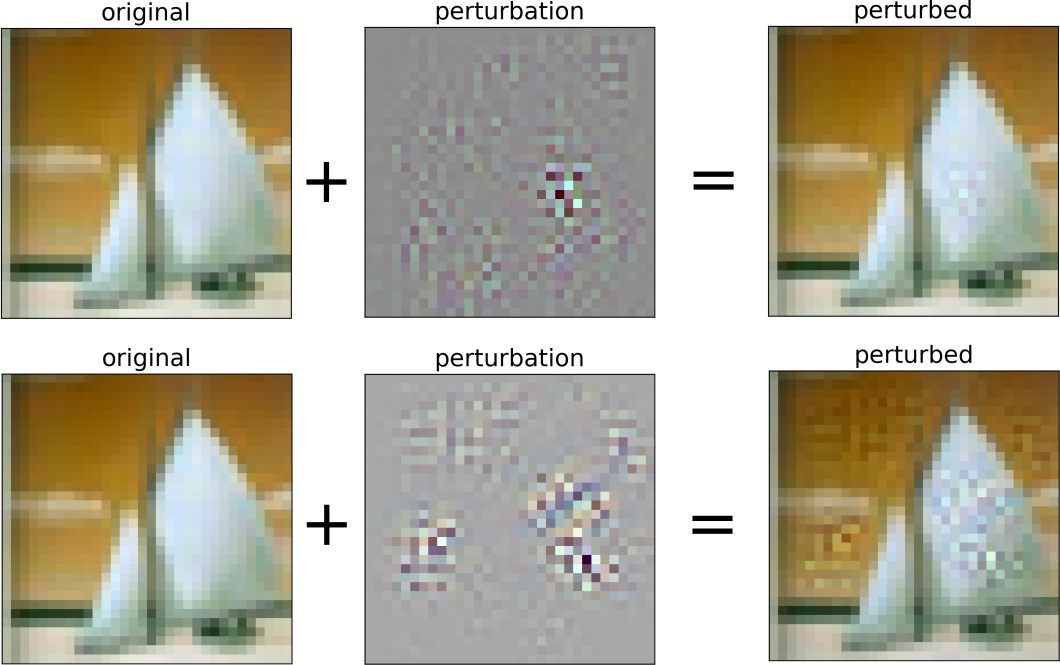}
\label{fig:targetedGrad}
}
\hspace{5mm}
\subfigure[Non-targeted (FGSM)]
{
\includegraphics[scale=0.225]{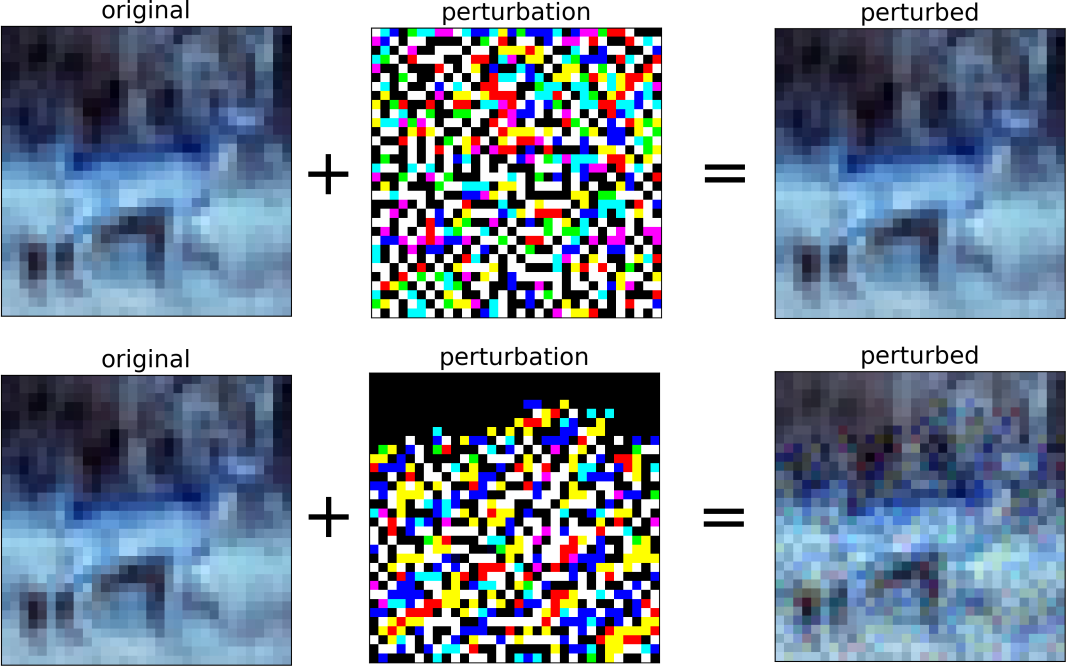}
\label{fig:nontargetedFGSM}
}
\caption{  Examples of targeted (a) and non-targeted (b) gradient-based perturbations generated with $\epsilon=0.1$. ResNet-32 is shown in the top row, whereas PR-ResNet-32 in the bottom row.  }
\label{fig:gradientPerturbations}
\end{figure}

\paragraph{Targeted attacks.}  
The above algorithm is applied to both ResNet-32 and PR-ResNet-32 on 100 test images 
from the CIFAR-10 dataset, sampled uniformly from all but the target class which is
selected as $\hat{y} = \textit{`cat'}$.
%
%
On ResNet-32, with $\epsilon=0.1$, the algorithm achieves a fooling rate 
of $16\%$ and $\mathbb{E}[ \vert\vert \mathbf{v} \vert\vert_2] = 125.59$, 
whereas PR-ResNet-32 with same configuration
is fooled only $3\%$ of the times and requires a much stronger attack with 
$\mathbb{E}\vert\vert \mathbf{v} \vert\vert_2 = 176.83$. An example of such perturbation is in Figure \ref{fig:targetedGrad}.

\paragraph{Non-targeted attacks.}  
Similarly to the targeted attacks, we sample 100 test images and, in this case
attempt to change the labels from $y$ to any other $\hat{y} \ne y$.
The algorithm is run with $\epsilon=0.1$ and fools ResNet-32 with a rate of 
$59\%$ and $\mathbb{E} \vert\vert \mathbf{v} \vert\vert_2 = 28.55$, almost three times
higher than that of PR-ResNet-32 that achieves a fooling rate of just 
$22\%$ with considerably larger 
$\mathbb{E} \vert\vert \mathbf{v} \vert\vert_2 = 102.11$.

%

\subsubsection{Fast Gradient Sign attacks}  
For the next series of experiments, we test against the 
{\em Fast Gradient Sign Method (FGSM)}~\cite{goodfellow2015}, 
top-ranked in the NIPS 2017 competition~\cite{kurakin2018};  
the implementation provided in Foolbox~\cite{rauber2017} is used here as well. 
%
FGSM generates an adversarial example $\hat{\mathbf{x}}$ 
for an input $\mathbf{x}$ with label $y$ as follows:
\begin{align*}
\hat{\mathbf{x}} & \leftarrow \mathbf{x} + \epsilon\, \mathrm{sign}(\nabla_{\mathbf{x}} J(\theta, \mathbf{x}, y)),
\end{align*}
where $J(\theta, \mathbf{x}, y)$ is the cost used during training
(cross-entropy in our case), and $\epsilon$ controls the magnitude 
of the perturbation. 
%
We test FGSM for both attacks, targeted and non-targeted.

\paragraph{Targeted attacks.}
To better compare with the algorithm of Section~\ref{sec:gradattack},  
we use the very same 100 test images and evaluation protocol.
FGSM scans ten $\epsilon$ values, from small to the one selected, 
and stops as soon as it succeeds.

Results are in line with what was reported for the gradient-based attack in Section~\ref{sec:gradattack} and
confirm the robustness of PeerNet to adversarial perturbations.
With $\epsilon=0.1$, ResNet-32 obtains a fooling rate of 
of $23\%$ and $\mathbb{E}\vert\vert \mathbf{v} \vert\vert_2 = 234.41$; for comparison, 
{PR-ResNet-32} yields a fooling rate of $8\%$ 
with $\mathbb{E}\vert\vert \mathbf{v} \vert\vert_2 = 416.36$.

\paragraph{Non-targeted attacks.} 
Also here, the same images and evaluation protocol were used as for the 
non-targeted gradient based attack, and FGSM tried to  fabricate a sample
that would change the predicted label into any of the other labels.
In this case, the values of $\epsilon = 0.01, 0.1, $ and $1.0$ were tested.
For ResNet-32, the fooling rates were $71.10\%, 89.70\%$, $98.90\%$ with 
$\mathbb{E}\vert\vert \mathbf{v} \vert\vert_2 =37.82, 109.41$, and $218.99$, respectively. 
Fooling rates for PR-ResNet-32 were, in this case, $17.70\%, 61.60\%$, and $91.21\%$ with 
$\mathbb{E}\vert\vert \mathbf{v} \vert\vert_2=32.64, 324.76$, and $608.98$, 
respectively, which reinforces the claim that PeerNets are an easy and effective
way to gain robustness against adversarial attacks. An example of such non-targeted perturbation is shown in Figure \ref{fig:nontargetedFGSM}. More examples of non-targeted perturbations are shown in Figure~\ref{fig:grad2}.

\subsection{Universal adversarial perturbations}
Finally, we used the method of~\cite{Dezfooli2016}, implemented in DeepFool 
toolbox \cite{Dezfooli2015} to generate a non-targeted 
universal adversarial perturbation $\mathbf{v}$.
The strength of the perturbation was bound by 
$\vert\vert \mathbf{v} \vert\vert_2 \leq \rho \mathbb{E}\vert\vert \mathbf{x} \vert\vert_2$, 
where the expectation is taken over the set of training images 
and the parameter $\rho$ controls the strength. 

\begin{figure}[ht!]
\centering
\addtocounter{subfigure}{-1}
\subfigure
{
\setlength\figureheight{4cm} 
\setlength\figurewidth{\linewidth}
\begin{tikzpicture}

\pgfplotsset{compat=newest} 

\definecolor{color0}{rgb}{1, 1, 1}

\tikzstyle{every node}=[font=\footnotesize]

\begin{axis}[
name=ax1,
at={(0\figurewidth,0\figureheight)},
xlabel={$\rho$},
ylabel={Fooling rate [\%]},
xmin=0.2, xmax=1.0,
ymin=0.0, ymax=100.0,
width=0.45\figurewidth,
height=\figureheight,
xmajorgrids,
x grid style={lightgray},
ymajorgrids,
y grid style={lightgray},
axis line style={black},
axis background/.style={fill=color0},
xtick={0.2, 0.4, 0.6, 0.8, 1.0},
xticklabels={0.2, 0.4, 0.6, 0.8, 1.0},
ytick={0.0, 20.0, 40.0, 60.0, 80.0, 100.00},
log ticks with fixed point,
legend columns=-1,
legend style={at={(0.6,-0.3)}, anchor=north west, font=\tiny, column sep=1ex},
legend cell align={left},
legend entries={{LeNet-5},{PR-LeNet-5}}
]

\addplot [line width=1.0pt, red]
table {%
0.2 7.1 
0.4 65.97
0.6 85.93
0.8 92.16
1.0 91.72
};
\addplot [line width=1.0pt, green]
table {%
0.2 4.56
0.4 5.97
0.6 11.67
0.8 46.43
1.0 50.09
};

\legend{}

\end{axis}

\begin{axis}[
at={(0.5\figurewidth,0\figureheight)},
xlabel={$\rho$},
ylabel={Accuracy [\%]},
xmin=0.2, xmax=1.0,
ymin=0.0, ymax=100.0,
width=0.45\figurewidth,
height=\figureheight,
xmajorgrids,
x grid style={lightgray},
ymajorgrids,
y grid style={lightgray},
axis line style={black},
axis background/.style={fill=color0},
xtick={0.2, 0.4, 0.6, 0.8, 1.0},
xticklabels={0.2, 0.4, 0.6, 0.8, 1.0},
ytick={0.0, 20.0, 40.0, 60.0, 80.0, 100.00},
log ticks with fixed point,
]

\addplot [line width=1.0pt, red]
table {%
0.2 92.68
0.4 33.87
0.6 14.11
0.8 7.88
1.0 8.24
};
\addplot [line width=1.0pt, green]
table {%
0.2 94.78
0.4 93.34
0.6 87.67
0.8 53.24
1.0 46.43
};
\end{axis}
\end{tikzpicture}  
\label{fig:foolMnistAll}
}
\caption{ Fooling rate and performance on MNIST dataset for different levels $\rho$ of universal adversarial noise. }
\label{fig:foolMnist}
\end{figure}
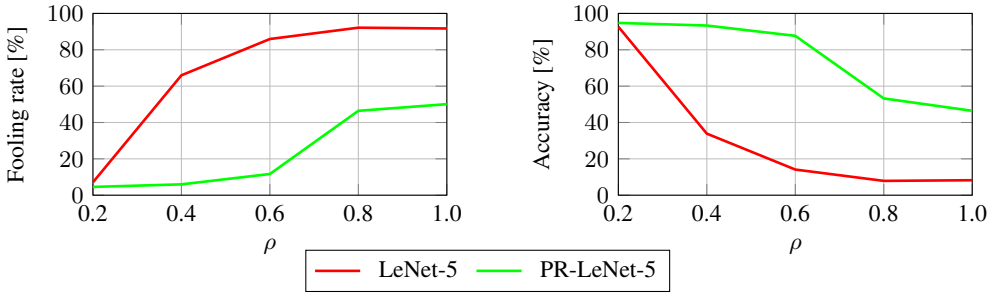

We first evaluated on the MNIST 
dataset~\cite{lecun1998mnist} and compared to LeNet-5 as baseline. 
Figure~\ref{fig:foolMnist} and Table \ref{tab:tableMnist} compare LeNet-5 and PR-LeNet-5 on MNIST dataset 
over increasing levels of perturbation $\rho = 0.2, 0.4, \hdots, 1.0$. 
Results clearly show that our PR-LeNet-5 variant is much more robust 
than its baseline counterpart LeNet-5. 

\begin{table}[ht!]
\begin{small}
\caption{Performance and fooling rates on the MNIST dataset for different levels $\rho$ of universal adversarial noise.}
\begin{center}
\begin{tabular}{lcccccc}
\toprule
\multirow{2}{1.5cm}{\centering\rule{0pt}{3mm} Method } & \multirow{2}{1.15cm}{\centering\rule{0pt}{4mm} Original Accuracy} & \multicolumn{5}{c}{Accuracy / Fooling Rate} \\
  \cmidrule{3-7}
&& $\rho=0.2$ & $\rho=0.4$ & $\rho=0.6$ & $\rho=0.8$ &  $\rho=1.0$ \\
\midrule
LeNet-5 & 98.6\% & 92.7\% / 7.1\% & 33.9\% / 66.0\% & 14.1\% / 85.9\% & 7.9\% / 92.2\% & 8.2\% / 91.7\% \\
PR-LeNet-5 & 98.2\% & 94.8\% / 4.6\% & 93.3\% / 6.0\% & 87.7\% / 11.7\% & 53.2\% / 46.4\% & 50.1\% / 50.1\% \\
\bottomrule
\label{tab:tableMnist}
\end{tabular}
\end{center}
\end{small}
\vspace*{-3mm}
\end{table}


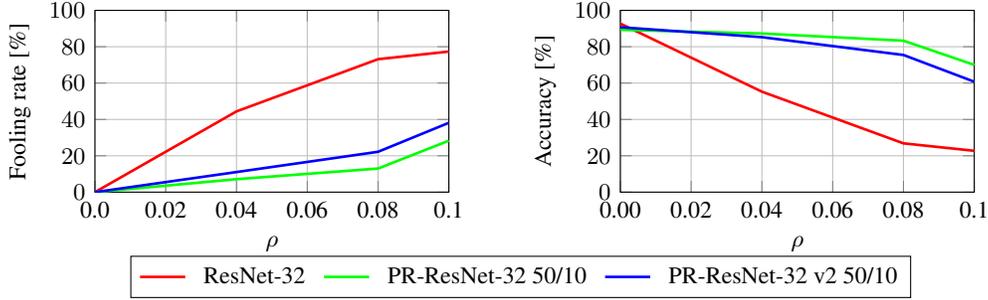
\begin{figure}[ht!]
\label{fig:cifarresults}
\centering
\subfigure
{
\setlength\figureheight{4cm} 
\setlength\figurewidth{\linewidth}
\begin{tikzpicture}

\pgfplotsset{compat=newest} 

\definecolor{color0}{rgb}{1, 1, 1}

\tikzstyle{every node}=[font=\footnotesize]

\begin{axis}[
name=ax1,
at={(0\figurewidth,0\figureheight)},
xlabel={$\rho$},
ylabel={Fooling rate [\%]},
xmin=0.0, xmax=0.1,
ymin=0.0, ymax=100.0,
width=0.45\figurewidth,
height=\figureheight,
xmajorgrids,
x grid style={lightgray},
ymajorgrids,
y grid style={lightgray},
axis line style={black},
axis background/.style={fill=color0},
xtick={0.0, 0.02, 0.04, 0.06, 0.08, 0.1},
xticklabels={0.0, 0.02, 0.04, 0.06, 0.08, 0.1},
ytick={0.0, 20.0, 40.0, 60.0, 80.0, 100.00},
log ticks with fixed point,
legend columns=-1,
legend style={at={(0.1,-0.35)}, anchor=north west, font=\tiny, column sep=1ex},
legend cell align={left},
legend entries={{ResNet-32},{PR-ResNet-32 50/10},{PR-ResNet-32 v2 50/10}}
]      

\addplot [line width=1.0pt, red]
table {%
0.00 0.00
0.04 44.42
0.08 73.14
0.10 77.34
};
\addplot [line width=1.0pt, green]
table {%
0.00 0.00
0.04 7.13
0.08 12.99
0.10 28.31
};
\addplot [line width=1.0pt, blue]
table {%
0.00 0.00
0.04 11.05
0.08 22.20
0.10 38.14
};

\legend{}

\end{axis}

\begin{axis}[
at={(0.5\figurewidth,0\figureheight)},
xlabel={$\rho$},
ylabel={Accuracy [\%]},
xmin=0.00, xmax=0.1,
ymin=0.0, ymax=100.0,
width=0.45\figurewidth,
height=\figureheight,
xmajorgrids,
x grid style={lightgray},
ymajorgrids,
y grid style={lightgray},
axis line style={black},
axis background/.style={fill=color0},
xtick={0.00, 0.02, 0.04, 0.06, 0.08, 0.1},
xticklabels={0.00, 0.02, 0.04, 0.06, 0.08, 0.1},
ytick={0.0, 20.0, 40.0, 60.0, 80.0, 100.00},
log ticks with fixed point,
]

\addplot [line width=1.0pt, red]
table {%
0.00 92.73
0.04 55.27
0.08 26.84
0.10 22.74
};
\addplot [line width=1.0pt, green]
table {%
0.00 89.30
0.04 87.27
0.08 83.32
0.10 70.01
};

\addplot [line width=1.0pt, blue]
table {%
0.00 90.72
0.04 85.26
0.08 75.46
0.10 60.75
};

\end{axis}
\end{tikzpicture}  
}
\caption{ Fooling rate and performance on CIFAR-10 dataset for different levels $\rho$ of universal adversarial noise. The legend for PR-Nets is in format PR-ResNet-32 $X$/$Y$, where $X$ is the graph size and $Y$ is the number of Monte Carlo runs.}
\label{fig:foolCifar10}
\end{figure}

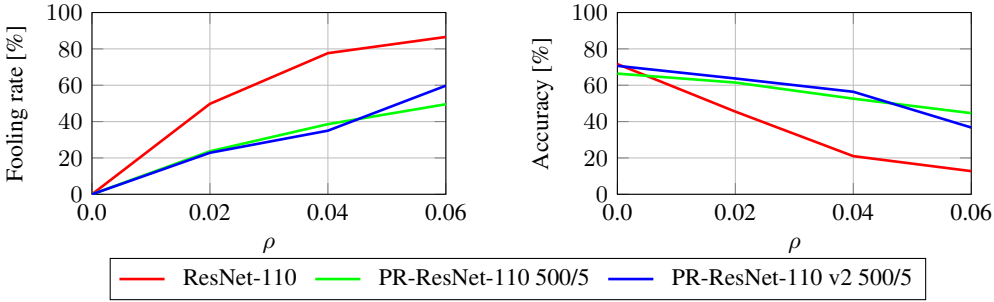
\begin{figure}[ht!]
\centering
\addtocounter{subfigure}{-1}
\subfigure
{
\setlength\figureheight{4cm} 
\setlength\figurewidth{\linewidth}
\begin{tikzpicture}

\pgfplotsset{compat=newest} 

\definecolor{color0}{rgb}{1, 1, 1}

\tikzstyle{every node}=[font=\footnotesize]

\begin{axis}[
name=ax1,
at={(0\figurewidth,0\figureheight)},
xlabel={$\rho$},
ylabel={Fooling rate [\%]},
xmin=0.0, xmax=0.06,
ymin=0.0, ymax=100.0,
width=0.45\figurewidth,
height=\figureheight,
xmajorgrids,
x grid style={lightgray},
ymajorgrids,
y grid style={lightgray},
axis line style={black},
axis background/.style={fill=color0},
xtick={0.0, 0.02, 0.04, 0.06},
xticklabels={0.0, 0.02, 0.04, 0.06},
ytick={0.0, 20.0, 40.0, 60.0, 80.0, 100.00},
scaled x ticks=false,
scaled y ticks=false,
log ticks with fixed point,
legend columns=-1,
legend style={at={(0.05,-0.35)}, anchor=north west, font=\tiny, column sep=1ex},
legend cell align={left},
legend entries={{ResNet-110},{PR-ResNet-110 500/5},{PR-ResNet-110 v2 500/5}}
]     

\addplot [line width=1.0pt, red]
table {%
0.00 0.00
0.02 49.78
0.04 77.64
0.06 86.56
};
\addplot [line width=1.0pt, green]
table {%
0.00 0.00
0.02 23.65
0.04 38.59
0.06 49.54
};
\addplot [line width=1.0pt, blue]
table {%
0.00 0.00
0.02 22.84
0.04 35.01
0.06 59.76 
};

\legend{}

\end{axis}

\begin{axis}[
at={(0.5\figurewidth,0\figureheight)},
xlabel={$\rho$},
ylabel={Accuracy [\%]},
xmin=0.0, xmax=0.06,
ymin=0.0, ymax=100.0,
width=0.45\figurewidth,
height=\figureheight,
xmajorgrids,
x grid style={lightgray},
ymajorgrids,
y grid style={lightgray},
axis line style={black},
axis background/.style={fill=color0},
xtick={0.0, 0.02, 0.04, 0.06},
xticklabels={0.0, 0.02, 0.04, 0.06},
ytick={0.0, 20.0, 40.0, 60.0, 80.0, 100.00},
scaled x ticks=false,
scaled y ticks=false,
log ticks with fixed point,
]

\addplot [line width=1.0pt, red]
table {%
0.00 71.63
0.02 45.49
0.04 20.99
0.06 12.74
};
\addplot [line width=1.0pt, green]
table {%
0.00 66.40
0.02 61.47 
0.04 52.61
0.06 44.64
};
\addplot [line width=1.0pt, blue]
table {%
0.00 70.66
0.02 63.71 
0.04 56.40
0.06 36.74
};

\end{axis}
\end{tikzpicture}  
}
\caption{ Fooling rate and performance on CIFAR-100 dataset for different levels $\rho$ of universal adversarial noise. The legend for PR-Nets is in format PR-ResNet-110 $X$/$Y$, where $X$ is the graph size and $Y$ is the number of Monte Carlo runs. }
\label{fig:foolCifar100}
\end{figure}

The same set of evaluations was performed on the CIFAR-10 and CIFAR-100 datasets, 
using ResNet-32 and ResNet-110 as baselines, respectively. 
The level of noise varied from $\rho = 0.02$ to $0.1$. 
Figures~\ref{fig:foolCifar10} and~\ref{fig:foolCifar100} summarize 
the results for the two models, confirming that our PR-ResNet variants exhibit 
far better resistance to universal adversarial attacks than the baselines in this setting as well. More detailed results are listed in Tables~\ref{tab:tableCifar10} and~\ref{tab:tableCifar100}.
The minor loss in accuracy at $\rho = 0$ is a consequence of the regularization due 
to the averaging in feature space, and we argue that it could be mitigated by 
increasing the model capacity. For this purpose, the same configuration with double 
number of maps in PR layers (marked as v2) was trained, and results show indeed a
sensible improvement 
that does not come at the cost of a much higher fooling rate 
as for the ResNet baselines.



\begin{table}[ht!]
\begin{small}
\caption{Performance and fooling rates on the CIFAR-10 dataset. ResNet-32 v2 and PR-ResNet-32 v2 have double the amount of feature maps after the last two convolutional blocks, meaning instead of (16, 16, 32, 64), it has (16, 16, 64, 128).}
\begin{center}
\begin{tabular}{ccccccc}
\toprule
\multirow{2}{1.5cm}{\centering\rule{0pt}{4mm} Method } & \multirow{2}{1.5cm}{\centering\rule{0pt}{4mm} Graph size } & \multirow{2}{1.5cm}{\centering\rule{0pt}{4mm} MC runs} & \multirow{2}{1.5cm}{\centering\rule{0pt}{4mm} Acc. orig [\%]} & \multicolumn{3}{c}{Acc pert. [\%] / Fool rate [\%]} \\
  \cmidrule{5-7}
&&&& $\rho=0.04$ & $\rho=0.08$ &  $\rho=0.10$ \\
\midrule
ResNet-32 & N/A & N/A & 92.73 & 55.27 / 44.42 & 26.84 / 73.14 & 22.74 / 77.34   \\
ResNet-32 v2 & N/A & N/A & 94.17 & 44.51 / 55.32 & 16.65 / 83.40 & 12.58 / 87.58   \\
PR-ResNet-32 & 50 & 1 & 88.18 & 87.27 / 7.98 & 82.43 / 14.08 & 69.33 / 28.80  \\
PR-ResNet-32 & 50 & 10 & 89.30 & 87.27 / 7.13 & 83.32 / 12.99 & 70.01 / 28.31 \\
PR-ResNet-32 & 100 & 5 & 89.19 & 87.33 / 7.43 & 83.37 / 13.20 & 70.11 / 28.19 \\
PR-ResNet-32 v2 & 50 & 10 & 90.72 & 85.26 / 11.05 & 75.46 / 22.20 & 60.75 / 38.14 \\
PR-ResNet-32 v2 & 100 & 5 & 90.65 & 85.35 / 11.25 & 75.94 / 21.82 & 61.10 / 37.77 \\
\bottomrule
\label{tab:tableCifar10}
\end{tabular}
\end{center}
\end{small}
\vspace*{-3mm}
\end{table}


\begin{table}[ht!]
\begin{small}
\caption{Performance and fooling rates on the CIFAR-100 dataset. PR-ResNet-110 v2 has double the amount of feature maps after the last two convolutional blocks, meaning instead of (16, 16, 32, 64), it has (16, 16, 64, 128).}
\begin{center}
\begin{tabular}{ccccccc}
\toprule
\multirow{2}{1.5cm}{\centering\rule{0pt}{4mm} Method } & \multirow{2}{1.5cm}{\centering\rule{0pt}{4mm} Graph size } & \multirow{2}{1.5cm}{\centering\rule{0pt}{4mm} MC runs} & \multirow{2}{1.5cm}{\centering\rule{0pt}{4mm} Acc. orig [\%]} & \multicolumn{3}{c}{Acc pert. [\%] / Fool rate [\%]} \\
  \cmidrule{5-7}
&&&& $\rho=0.02$ & $\rho=0.04$ &  $\rho=0.06$ \\
\midrule
ResNet-110 & N/A & N/A & 71.63 & 45.49 / 49.78 & 20.99 / 77.64 & 12.74 / 86.56  \\
PR-ResNet-110 & 500 & 5 & 66.40 & 61.47 / 23.65 & 52.61 / 38.59 & 44.64 / 49.54\\ 
PR-ResNet-110 v2 & 500 & 5 & 70.66 & 63.71 / 22.84 & 56.40 / 35.01 & 36.74 / 59.76 \\
\bottomrule
\label{tab:tableCifar100}
\end{tabular}
\end{center}
\end{small}
\vspace*{-3mm}
\end{table}


A visual comparison of perturbations generated for classical CNN and our PR-Net 
is depicted in Figure~\ref{fig:perturbationUniversal}. More examples can be found in Figure~\ref{fig:universal5}. 
It is apparent that the perturbations for PR-Net have more localized structures. We argue that this is due
to the attempt of fooling the KNN mechanism of the PR layers, resulting in strong noise in `background' areas rather than in the central parts usually containing the object.

\section{Conclusions}
We introduced PeerNets, a novel family of deep networks for image understanding, alternating Euclidean and Graph convolutions 
to harness information from peer images. We showed the robustness of PeerNets to adversarial attacks 
in a variety of scenarios through extensive experiments for white-box attacks 
in targeted and non-targeted settings.
PeerNets are simple to use, can be added to any baseline model with minimal changes, and 
are able to deliver remarkably lower fooling rates with negligible loss in performance. 
Interestingly, the amount of noise required to fool PeerNets is much higher and results in 
the generation of new images where the noise has clear structure and is significantly more perceivable to the human eye. 
In future work, we plan to provide a theoretical analysis of the method
and scale it to ImageNet-like benchmarks.

\subsubsection*{Acknowledgments}
JS, FM and MB are supported in part by ERC Consolidator Grant No. 724228 (LEMAN), Google Faculty Research Awards, an Amazon AWS Machine Learning Research grant, an Nvidia equipment grant, a Radcliffe Fellowship at the Institute for Advanced Study, Harvard University, and a Rudolf Diesel Industrial Fellowship at IAS TU Munich. LG is supported by NSF grants DMS-1546206 and DMS-1521608, a DoD Vannevar Bush Faculty Fellowship, an Amazon AWS Machine Learning Research grant, a Google Focused Research Award, and a Hans Fischer Fellowship at IAS TU Munich.


\bibliographystyle{plain}
\bibliography{sample}

\begin{thebibliography}{10}

\bibitem{atwood2016diffusion}
J.~Atwood and D.~Towsley.
\newblock Diffusion-convolutional neural networks.
\newblock In {\em Proc. NIPS}, 2016.

\bibitem{badrinarayanan2015segnet}
V.~Badrinarayanan, A.~Kendall, and R.~Cipolla.
\newblock {SegNet}: A deep convolutional encoder-decoder architecture for image
  segmentation.
\newblock {\em IEEE Trans. PAMI}, 2017.

\bibitem{belkin2003laplacian}
M.~Belkin and P.~Niyogi.
\newblock Laplacian eigenmaps for dimensionality reduction and data
  representation.
\newblock {\em Neural Computation}, 15(6):1373--1396, 2003.

\bibitem{bronstein2017geometric}
M.~M. Bronstein, J.~Bruna, Y.~LeCun, A.~Szlam, and P.~Vandergheynst.
\newblock Geometric deep learning: going beyond euclidean data.
\newblock {\em IEEE Signal Process. Mag.}, 34(4):18--42, 2017.

\bibitem{bruna2013spectral}
J.~Bruna, W.~Zaremba, A.~Szlam, and Y.~LeCun.
\newblock Spectral networks and locally connected networks on graphs.
\newblock {\em arXiv:1312.6203}, 2013.

\bibitem{Buades2005}
A.~Buades, B.~Coll, and J.-M. Morel.
\newblock A non-local algorithm for image denoising.
\newblock In {\em Proc. CVPR}, 2005.

\bibitem{coifman2006diffusion}
R.~R Coifman and S.~Lafon.
\newblock Diffusion maps.
\newblock {\em Applied and Computational Harmonic Analysis}, 21(1):5--30, 2006.

\bibitem{defferrard2016convolutional}
M.~Defferrard, X.~Bresson, and P.~Vandergheynst.
\newblock Convolutional neural networks on graphs with fast localized spectral
  filtering.
\newblock In {\em Proc. NIPS}, 2016.

\bibitem{duvenaud2015convolutional}
D.~K Duvenaud, D.~Maclaurin, J.~Iparraguirre, R.~Bombarell, T.~Hirzel,
  A.~Aspuru-Guzik, and R.~P Adams.
\newblock Convolutional networks on graphs for learning molecular fingerprints.
\newblock In {\em Proc. NIPS}, 2015.

\bibitem{fawzi2018analysis}
A.~Fawzi, O.~Fawzi, and P.~Frossard.
\newblock Analysis of classifiers’ robustness to adversarial perturbations.
\newblock {\em Machine Learning}, 107(3):481--508, 2018.

\bibitem{garcia2017few}
V.~Garcia and J.~Bruna.
\newblock Few-shot learning with graph neural networks.
\newblock {\em arXiv:1711.04043}, 2017.

\bibitem{gilmer2017neural}
J.~Gilmer, S.~S Schoenholz, P.~F Riley, O.~Vinyals, and G.~E Dahl.
\newblock Neural message passing for quantum chemistry.
\newblock {\em arXiv:1704.01212}, 2017.

\bibitem{goodfellow2015}
I.~Goodfellow, J.~Shlens, and Ch. Szegedy.
\newblock Explaining and harnessing adversarial examples.
\newblock In {\em Proc. ICLR}, 2015.

\bibitem{gori2005new}
M.~Gori, G.~Monfardini, and F.~Scarselli.
\newblock A new model for learning in graph domains.
\newblock In {\em Proc. IJCNN}, 2005.

\bibitem{gu2014towards}
S.~Gu and L.~Rigazio.
\newblock Towards deep neural network architectures robust to adversarial
  examples.
\newblock {\em arXiv:1412.5068}, 2014.

\bibitem{hamilton2017inductive}
W.~Hamilton, Z.~Ying, and J.~Leskovec.
\newblock Inductive representation learning on large graphs.
\newblock In {\em Proc. NIPS}, 2017.

\bibitem{he2017maskrcnn}
K.~He, G.~Gkioxari, P.~Doll\'{a}r, and R.~Girshick.
\newblock {Mask R-CNN}.
\newblock In {\em Proc. ICCV}, 2017.

\bibitem{he2015b}
K.~He, X.~Zhang, S.~Ren, and J.~Sun.
\newblock Deep residual learning for image recognition.
\newblock In {\em Proc. ICCV}, 2015.

\bibitem{henaff2015deep}
M.~Henaff, J.~Bruna, and Y.~LeCun.
\newblock Deep convolutional networks on graph-structured data.
\newblock {\em arXiv:1506.05163}, 2015.

\bibitem{kipf2016semi}
T.~N Kipf and M.~Welling.
\newblock Semi-supervised classification with graph convolutional networks.
\newblock {\em arXiv:1609.02907}, 2016.

\bibitem{krizhevsky2009learning}
A.~Krizhevsky.
\newblock Learning multiple layers of features from tiny images.
\newblock 2009.

\bibitem{krizhevsky2012a}
A.~Krizhevsky, I.~Sutskever, and G.~E. Hinton.
\newblock {ImageNet} classification with deep convolutional neural networks.
\newblock In {\em Proc. NIPS}, 2012.

\bibitem{kurakin2018}
A.~Kurakin, I.~J. Goodfellow, S.~Bengio, Y.~Dong, F.~Liao, M.~Liang, T.~Pang,
  J.~Zhu, X.~Hu, C.~Xie, J.~Wang, Z.~Zhang, Z.~Ren, A.~L. Yuille, S.~Huang,
  Y.~Zhao, Y.~Zhao, Z.~Han, J.~Long, Y.~Berdibekov, T.~Akiba, S.~Tokui, and
  M.~Abe.
\newblock Adversarial attacks and defences competition.
\newblock {\em arXiv:1804.00097}, 2018.

\bibitem{lecun1998mnist}
Y.~LeCun.
\newblock The {MNIST} database of handwritten digits.
\newblock {\em http://yann.lecun.com/exdb/mnist/}, 1998.

\bibitem{lecun1998gradient}
Y.~LeCun, L.~Bottou, Y.~Bengio, and P.~Haffner.
\newblock Gradient-based learning applied to document recognition.
\newblock {\em Proc. IEEE}, 86(11):2278--2324, 1998.

\bibitem{levie2017cayleynets}
R.~Levie, F.~Monti, X.~Bresson, and M.~M Bronstein.
\newblock Cayleynets: Graph convolutional neural networks with complex rational
  spectral filters.
\newblock {\em arXiv:1705.07664}, 2017.

\bibitem{li2016gated}
Y.~Li, D.~Tarlow, M.~Brockschmidt, and R.~Zemel.
\newblock Gated graph sequence neural networks.
\newblock In {\em Proc. ICLR}, 2016.

\bibitem{liuYTLA17}
Z.~Liu, R.~A. Yeh, X.~Tang, Y.~Liu, and A.~Agarwala.
\newblock Video frame synthesis using deep voxel flow.
\newblock In {\em Proc. ICCV}, 2017.

\bibitem{mahdizadehaghdam2018deep}
S.~Mahdizadehaghdam, A.~Panahi, H.~Krim, and L.~Dai.
\newblock Deep dictionary learning: A parametric network approach.
\newblock {\em arXiv:1803.04022}, 2018.

\bibitem{monti2016geometric}
F.~Monti, D.~Boscaini, J.~Masci, E.~Rodol{\`a}, J.~Svoboda, and M.~M.
  Bronstein.
\newblock Geometric deep learning on graphs and manifolds using mixture model
  {CNN}s.
\newblock In {\em Proc. CVPR}, 2017.

\bibitem{monti2017geometric}
F.~Monti, M.~M. Bronstein, and X.~Bresson.
\newblock Geometric matrix completion with recurrent multi-graph neural
  networks.
\newblock In {\em Proc. NIPS}, 2017.

\bibitem{monti2018motifnet}
F.~Monti, K.~Otness, and M.~M Bronstein.
\newblock Motifnet: a motif-based graph convolutional network for directed
  graphs.
\newblock {\em arXiv:1802.01572}, 2018.

\bibitem{Dezfooli2016}
S.{-}M. Moosavi{-}Dezfooli, A.~Fawzi, O.~Fawzi, and P.~Frossard.
\newblock Universal adversarial perturbations.
\newblock {\em arXiv:1610.08401}, 2016.

\bibitem{Dezfooli2015}
S.{-}M. Moosavi{-}Dezfooli, A.~Fawzi, and P.~Frossard.
\newblock Deepfool: a simple and accurate method to fool deep neural networks.
\newblock {\em arXiv:1511.04599}, 2015.

\bibitem{niklausML17}
S.~Niklaus, L.~Mai, and F.~Liu.
\newblock Video frame interpolation via adaptive convolution.
\newblock In {\em Proc. {CVPR}}, 2017.

\bibitem{papernot2016effectiveness}
N.~Papernot and P.~McDaniel.
\newblock On the effectiveness of defensive distillation.
\newblock {\em arXiv:1607.05113}, 2016.

\bibitem{rauber2017}
J.~Rauber, W.~Brendel, and M.~Bethge.
\newblock Foolbox v0.8.0: {A} python toolbox to benchmark the robustness of
  machine learning models.
\newblock {\em arXiv:1707.04131}, 2017.

\bibitem{redmonDGF16}
J.~Redmon, S.~Kumar Divvala, R.~B. Girshick, and A.~Farhadi.
\newblock You only look once: Unified, real-time object detection.
\newblock In {\em Proc. CVPR}, 2016.

\bibitem{renNIPS15fasterrcnn}
S.~Ren, K.~He, R.~Girshick, and J.~Sun.
\newblock Faster {R-CNN}: Towards real-time object detection with region
  proposal networks.
\newblock In {\em Proc. NIPS}, 2015.

\bibitem{scarselli2009graph}
F.~Scarselli, M.~Gori, Ah~C. Tsoi, M.~Hagenbuchner, and G.~Monfardini.
\newblock The graph neural network model.
\newblock {\em IEEE Trans. Neural Netw.}, 20(1):61--80, 2009.

\bibitem{sharif2016accessorize}
M.~Sharif, S.~Bhagavatula, L.~Bauer, and M.~K Reiter.
\newblock Accessorize to a crime: Real and stealthy attacks on state-of-the-art
  face recognition.
\newblock In {\em Proc. CCS}, 2016.

\bibitem{sochen1998general}
N.~Sochen, R.~Kimmel, and R.~Malladi.
\newblock A general framework for low level vision.
\newblock {\em IEEE Trans. Image Proc.}, 7(3):310--318, 1998.

\bibitem{suOnePixAttack2017}
J.~Su, D.~Vasconcellos Vargas, and K.~Sakurai.
\newblock One pixel attack for fooling deep neural networks.
\newblock {\em arXiv:1710.08864}, 2017.

\bibitem{szegedy2013intriguing}
Ch. Szegedy, W.~Zaremba, I.~Sutskever, J.~Bruna, D.~Erhan, I.~Goodfellow, and
  R.~Fergus.
\newblock Intriguing properties of neural networks.
\newblock {\em arXiv:1312.6199}, 2013.

\bibitem{tomasi1998bilateral}
C.~Tomasi and R.~Manduchi.
\newblock Bilateral filtering for gray and color images.
\newblock In {\em Proc. ICCV}, 1998.

\bibitem{tramer2017ensemble}
F.~Tram{\`e}r, A.~Kurakin, N.~Papernot, D.~Boneh, and P.~McDaniel.
\newblock Ensemble adversarial training: Attacks and defenses.
\newblock {\em arXiv:1705.07204}, 2017.

\bibitem{tramer2018ensemble}
F.~Tramèr, A.~Kurakin, N.~Papernot, I.~Goodfellow, Dan Boneh, and P.~McDaniel.
\newblock Ensemble adversarial training: Attacks and defenses.
\newblock In {\em Proc. ICLR}, 2018.

\bibitem{velivckovic2017graph}
P.~Veli{\v{c}}kovi{\'c}, G.~Cucurull, A.~Casanova, A.~Romero, P.~Li{\`o}, and
  Y.~Bengio.
\newblock Graph attention networks.
\newblock {\em arXiv:1710.10903}, 2017.

\bibitem{vijayanarasimhan17}
S.~Vijayanarasimhan, S.~Ricco, C.~Schmid, R.~Sukthankar, and K.~Fragkiadaki.
\newblock {SfM-Net}: Learning of structure and motion from video.
\newblock {\em arXiv:1704.07804}, 2017.

\bibitem{wang2017non}
X.~Wang, R.~Girshick, A.~Gupta, and K.~He.
\newblock Non-local neural networks.
\newblock {\em arXiv:1711.07971}, 2017.

\bibitem{wang2018dynamic}
Y.~Wang, Y.~Sun, Z.~Liu, S.~E Sarma, M.~M Bronstein, and J.~M Solomon.
\newblock Dynamic graph {CNN} for learning on point clouds.
\newblock {\em arXiv:1801.07829}, 2018.

\bibitem{yuanLWYG17}
Y.~Yuan, X.~Liang, X.~Wang, D.{-}Y. Yeung, and A.~Gupta.
\newblock Temporal dynamic graph {LSTM} for action-driven video object
  detection.
\newblock In {\em Proc. {ICCV}}, 2017.

\bibitem{zhu2017ldmnet}
W.~Zhu, Q.~Qiu, J.~Huang, R.~Calderbank, G.~Sapiro, and I.~Daubechies.
\newblock Ldmnet: Low dimensional manifold regularized neural networks.
\newblock {\em arXiv:1711.06246}, 2017.

\end{thebibliography}

\newpage
\appendix






\label{app:UniversalAdvPertVisually}

\begin{figure}[ht]
\centering
\subfigure
{
\includegraphics[scale=0.234]{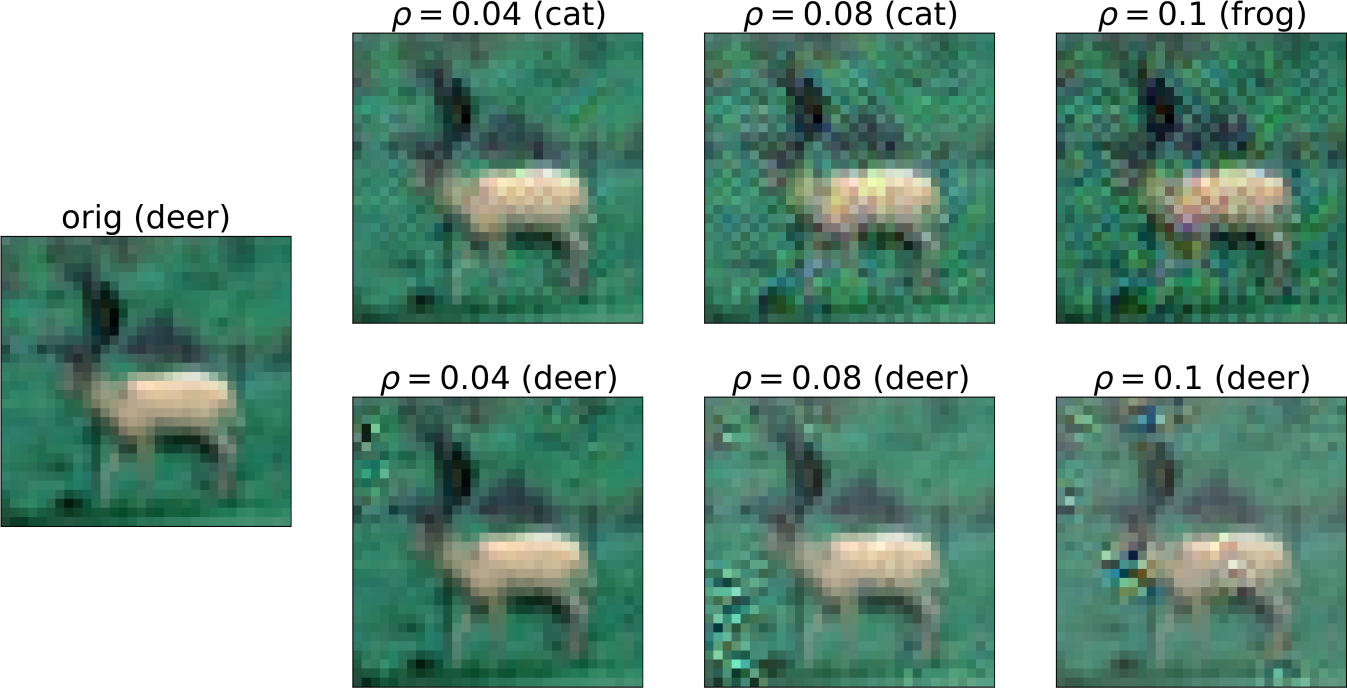}
\label{fig:universal2}
}
\subfigure
{
\includegraphics[scale=0.234]{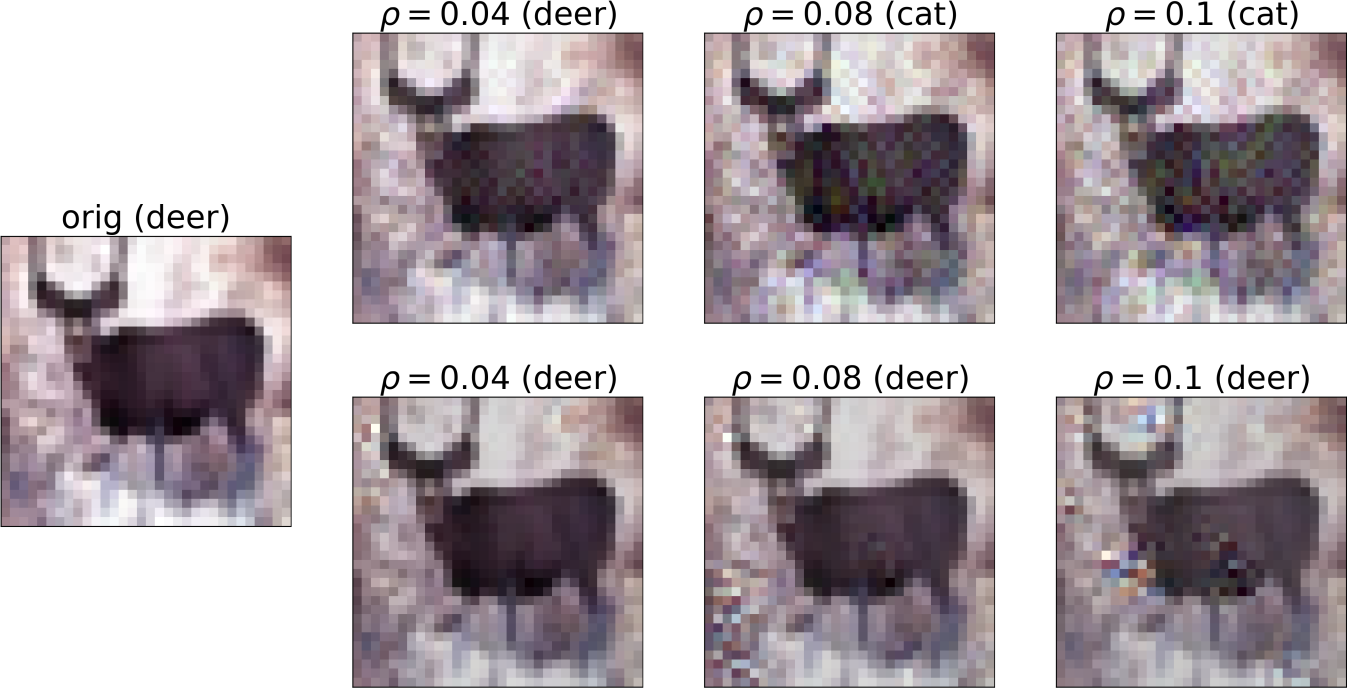}
\label{fig:universal3}
}
\subfigure
{
\includegraphics[scale=0.234]{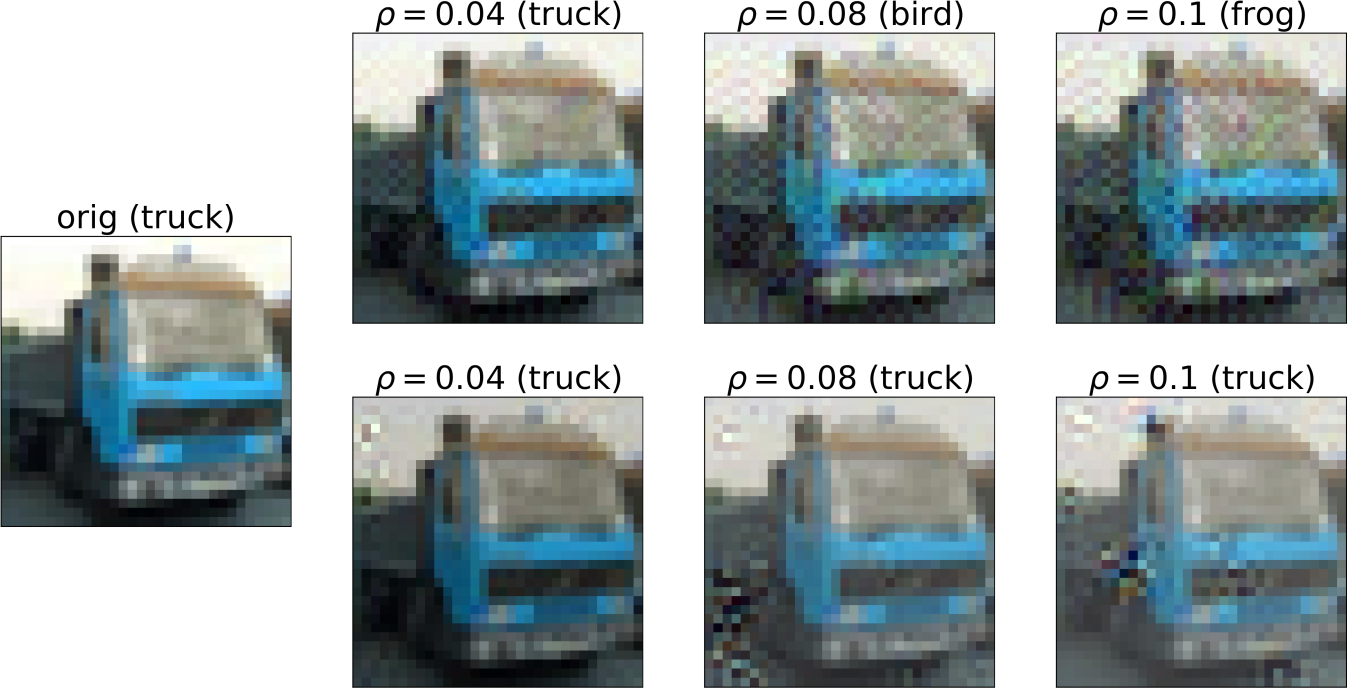}
\label{fig:universal4}
}
\subfigure
{
\includegraphics[scale=0.234]{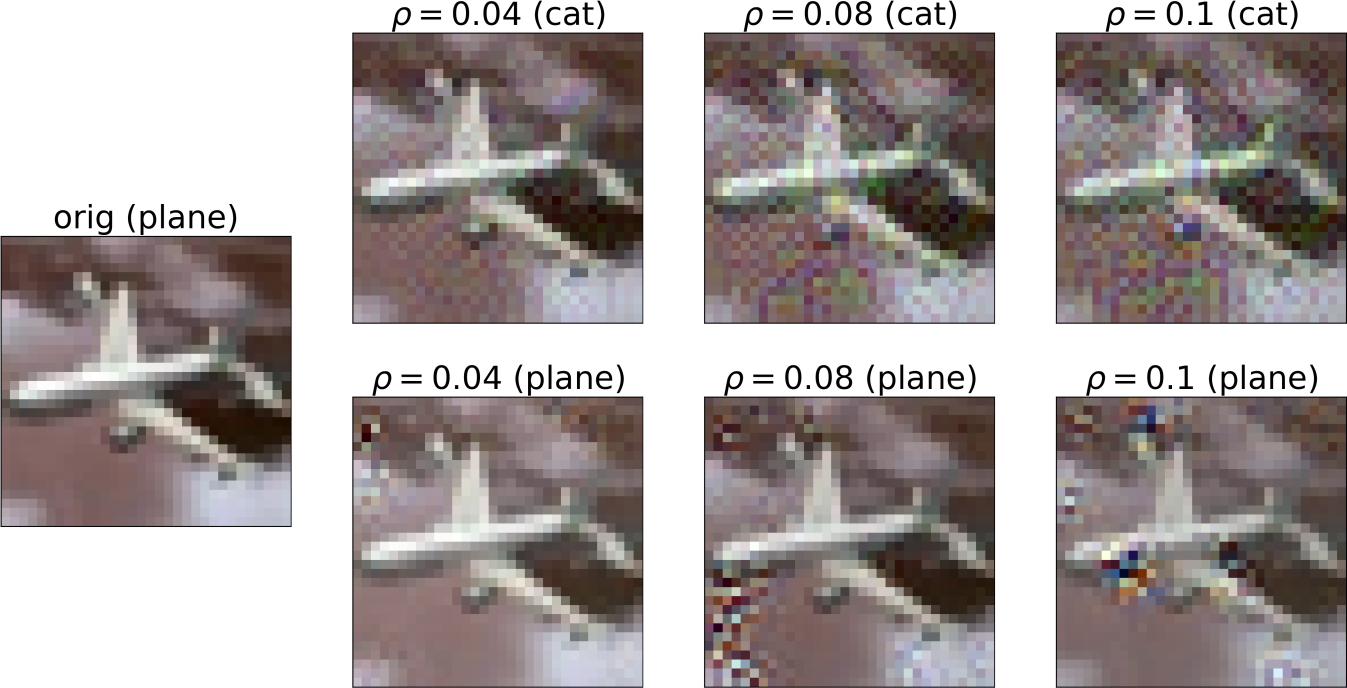}
\label{fig:universal5}
}
\caption{ Examples of applying universal perturbations to different images from the CIFAR-10 test set. For each sample, the leftmost is the original image, then adversarial examples for ResNet-32 are shown in the top row, whereas for PR-ResNet-32 in the bottom row. The number in the brackets is the predicted class of the image for comparison.  }
\label{fig:gradientPerturbations}
\end{figure}

\newpage

\label{app:GradientBasedVisually}

\begin{figure}[ht]
\centering
\subfigure[Non-targeted FGSM]
{
\includegraphics[scale=0.225]{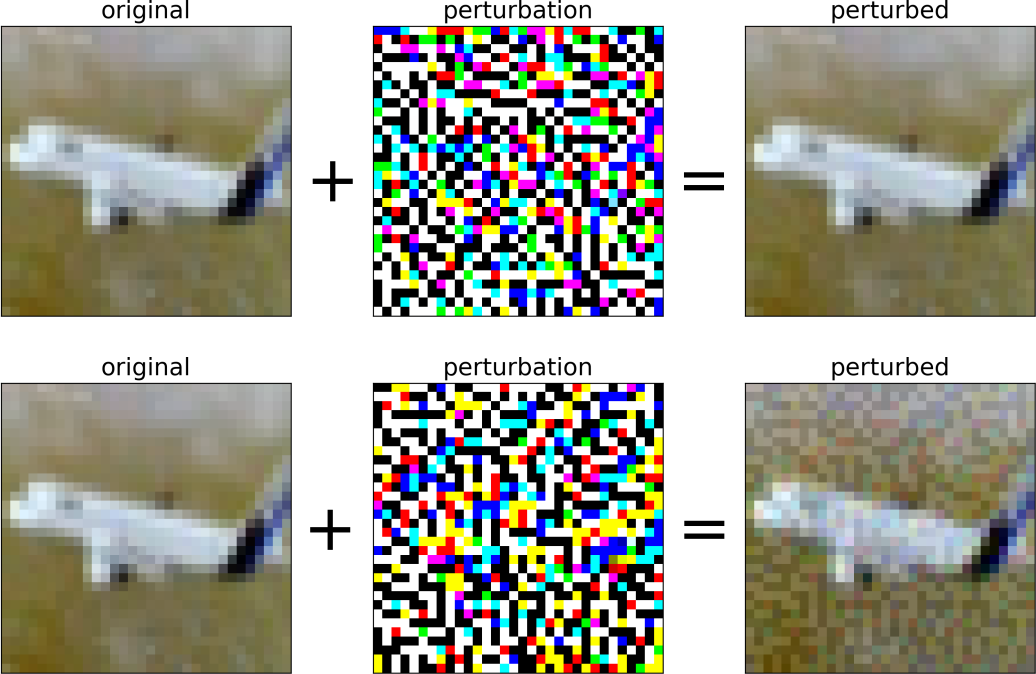}
\label{fig:fgsm1}
}
\vspace{2mm}
\hspace{5mm}
\subfigure[Non-targeted gradient descent]
{
\includegraphics[scale=0.225]{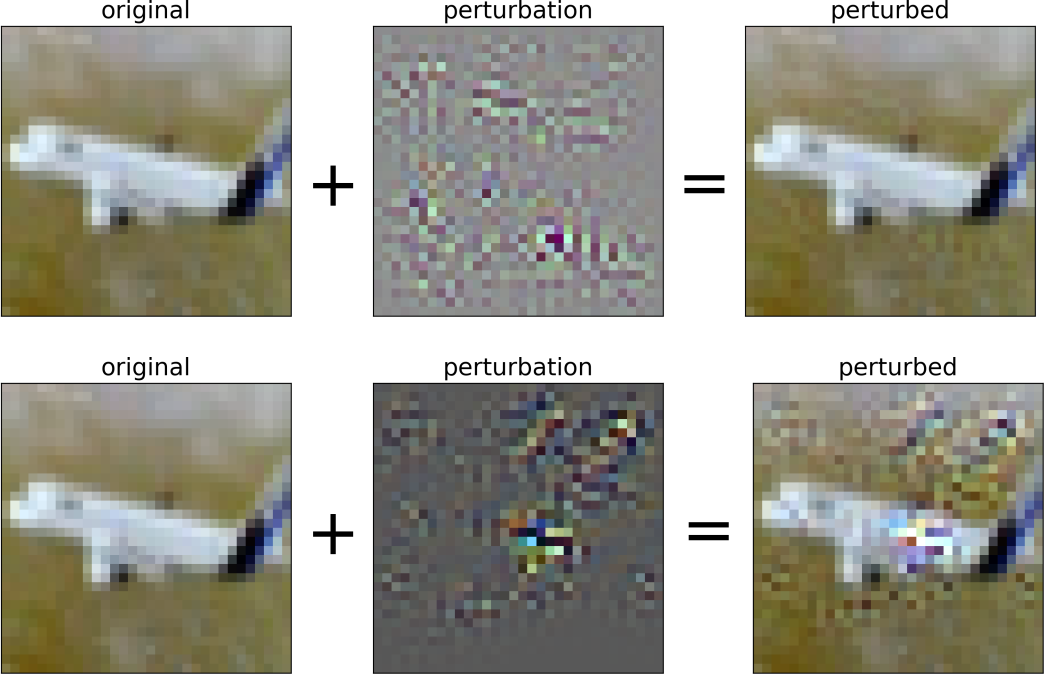}
\label{fig:fgsm2}
}
\subfigure[Non-targeted FGSM]
{
\includegraphics[scale=0.225]{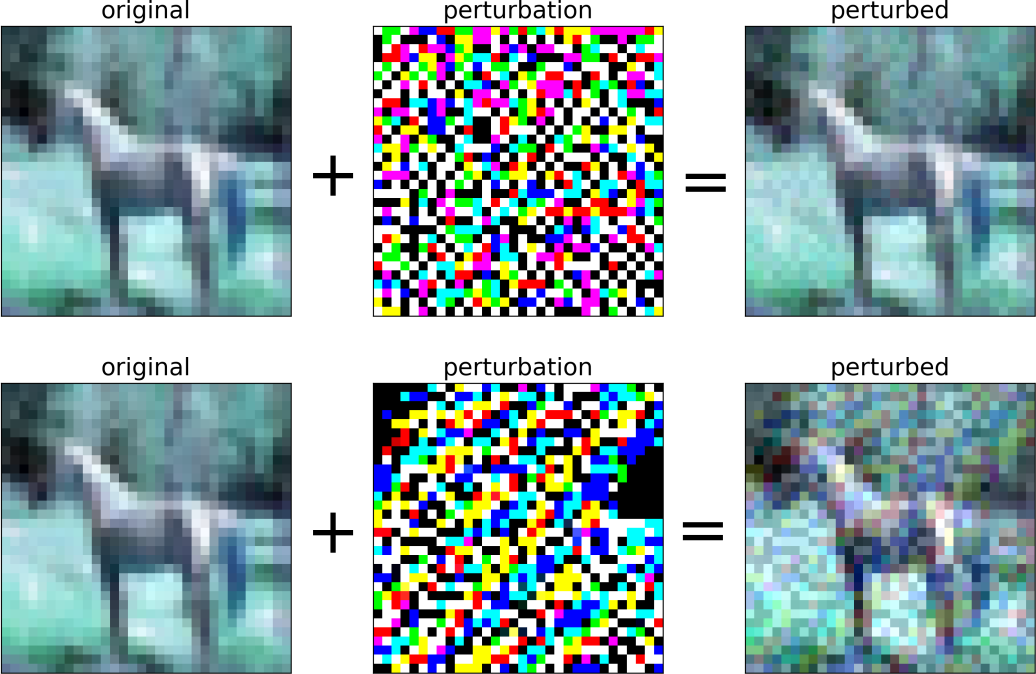}
\label{fig:grad1}
}
\vspace{2mm}
\hspace{5mm}
\subfigure[Non-targeted gradient descent]
{
\includegraphics[scale=0.225]{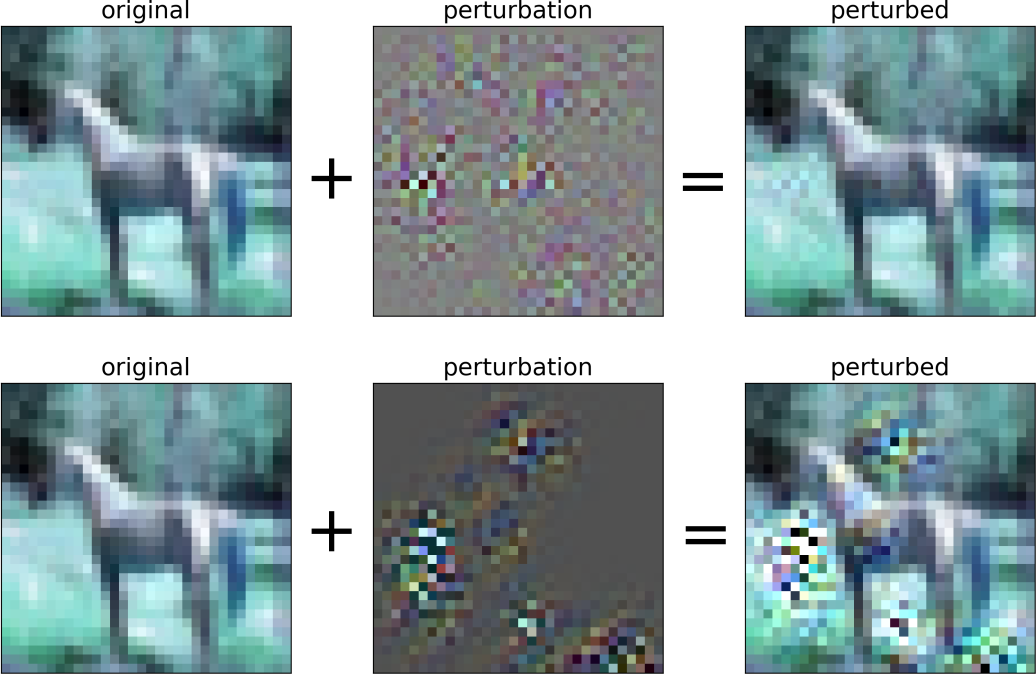}
\label{fig:grad2}
}
\caption{ Examples of non-targeted perturbations for different images from the CIFAR-10 test set using FGSM((a), (c)) and gradient descent((b), (d)). For each sample, results for ResNet-32 are shown in the top row, whereas for PR-ResNet-32 in the bottom row. In order to successfully generate adversarial samples for PR-Nets, the magnitude of the perturbation has to be much higher. }
\label{fig:gradientPerturbations}
\end{figure}

\end{document}